\documentclass[12pt]{article}
\usepackage[round]{natbib}
\usepackage[utf8]{inputenc}
\usepackage{graphicx}
\usepackage{caption}
\usepackage{subcaption}
\usepackage{float}
\usepackage{fullpage}
\usepackage{setspace}
\usepackage{amsmath,amssymb,bbm,amsthm,stackengine}
\newtheorem{theorem}{Theorem}
\newtheorem{lemma}{Lemma}
\newtheorem{proposition}{Proposition}
\newtheorem{assumption}{A\ignorespaces}
\usepackage{hyperref}
\newcommand{\R}{\mathbb{R}}
\renewcommand{\P}{\mathbb{P}}
\newcommand{\B}{\mathbb{B}}
\newcommand{\E}{\mathbb{E}}

\newcommand{\x}{\mathbf{x}}

\renewcommand{\u}{\mathbf{u}}
\renewcommand{\v}{\mathbf{v}}
\newcommand{\X}{\mathbf{X}}
\newcommand{\Rc}{\mathcal{R}}
\renewcommand{\L}{\mathcal{L}}

\newcommand{\1}[1]{\mathbbm{1}\left[{#1}\right]}
\DeclareMathOperator*{\argmin}{argmin}

\usepackage[ruled,vlined]{algorithm2e}
\usepackage[dvipsnames]{xcolor}

\title{Prediction in the presence of response-dependent missing labels}
\date{}
\author{Hyebin Song\thanks{Department of Statistics, The Pennsylvania State University}, Garvesh Raskutti\thanks{Department of Statistics, University of Wisconsin-Madison}, Rebecca Willett\thanks{Department of Statistics, University of Chicago} }

\begin{document}
\maketitle


\begin{abstract}
In a variety of settings, limitations of sensing technologies or other sampling mechanisms result in missing labels, where the likelihood of a missing label in the training set is an unknown function of the data. For example, satellites used to detect forest fires cannot sense fires below a certain size threshold. In such cases, training datasets consist of positive and pseudo-negative observations where pseudo-negative observations can be either true negatives or undetected positives with small magnitudes. We develop a new methodology and non-convex algorithm P(ositive) U(nlabeled) - O(ccurrence) M(agnitude) M(ixture) which jointly estimates the occurrence and detection likelihood of positive samples, utilizing prior knowledge of the detection mechanism. Our approach uses ideas from positive-unlabeled (PU)-learning and zero-inflated models that jointly estimate the magnitude and occurrence of events. We provide conditions under which our model is identifiable and prove that even though our approach leads to a non-convex objective, any local minimizer has optimal statistical error (up to a log term) and projected gradient descent has geometric convergence rates. We demonstrate on both synthetic data and a California wildfire dataset that our method out-performs existing state-of-the-art approaches.

\end{abstract}

\section{Introduction}
A common challenge in many statistical machine learning problems is \emph{noisy or missing labels.} In such settings, it is often common to assume the labels are missing at random and place a distribution on the missing labels (see e.g.~\citealp{Little2019-ib, molenberghs2014handbook}). However, in many applications, labels are missing systematically due to aspects of the technology in the data collection process.
Consider, for example, a dataset consisting of wildfire events in California where fire size is measured using satellite imagery. Due to the limited resolution of the satellite optics, fires smaller than a certain threshold will not be observed, complicating the effort of building a predictor of fire size. Similarly, consider forecasting the spread or impact of a virus, where a person's likelihood of being tested and included in a dataset depends on the severity of their symptoms. These are both examples of \emph{response-dependent missing labels} where labels or measurements are missing based on the \emph{magnitude or size} of the measured event. This response-dependent sampling bias poses a significant challenge in terms of (i) predicting event (such as fire) occurrence, since small magnitude events are not recorded and (ii) predicting the magnitude of each event (due to positive bias of the measurements).

In this paper, we develop a statistical framework that addresses \emph{response-dependent missing labels} with a two-level model that (i) models  the {\em true event magnitude} $Y$ as a mixture of $0$, indicating no event, and a positive distribution if the event occurs; and (ii) models the \emph{observed event magnitude} $Z$, which is either the same as $Y$ or $0$, depending on the  true response  $Y$. More specifically,
$$
P(Z = 0|Y = y > 0, \mathbf{X} = \x) = 1-\Gamma(y),
$$
where $\mathbf{X}=\x$ denotes the features or covariates and $\Gamma(y)$ represents a probability depending on $y$ which accounts for the outcome-dependence. Hence $Z=0$ could either denote a ``true'' negative where $Y=0$ or a ``false negative'' where $Y>0$ but $Z=0$.

This flexible framework allows us to model response-dependent missing labels through an occurrence-magnitude mixture distribution for $Y$ and the probability function $\Gamma(y)$ for the observed response $Z$. This model presents both identifiability and computational challenges that we address in this paper. Since $Z = 0$ could either denote a true $0$ or a false $0$, we first provide identifiablity conditions on our mixed model. Secondly, two computational challenges arise: (i) the likelihood of the observed data $Z$ involves integration over the function $\Gamma(y)$ and (ii) even if this integration is possible, the objective is non-convex. To address (i), we choose $\Gamma(y)$ -- the CDF of a Gamma distribution which allows a closed-form computation of the integral; to address (ii), we demonstrate that even though the objective is non-convex, using projected gradient descent leads to a local minimizer with desirable statistical properties.

\subsection{Related Work}

\noindent{\bf Censored labels:} Our proposed model is in contrast with the Type I Tobit model \citep{Tobin1958-uw}, where excess zeros arise due to the censoring of an underlying continuous variable. In the case of the Tobit model, zeros are only proxies for values below a certain thereshold, and therefore the goal of Tobit analysis is to estimate magnitude only. On the contrary, our framework models the two‐part mixture models that separately model the probability of event occurrences and magnitude of the events \citep{Smith2014-xf, Neelon2016-vj}.

\vspace{1em}
\noindent{\bf Positive-Unlabeled (PU) and multi-label learning:} Our approach is also related to Positive-Unlabeled (PU) and multi-label learning.
 PU-learning is a kind of semi-supervised learning where learning is performed based on positive and unlabeled examples \citep{Liu2003-hv, Elkan2008-aq,Du_Plessis2015-lo}. In particular, response labels are only partially known, as unlabeled examples can belong to either the positive or negative class. Similarly, in multi-label learning problems, each example is associated with only a subset of the true relevant labels \citep{Jain2016-ks, Schultheis2020-ru}.  In both cases, the goal is to learn a model that can predict the occurrence of true labels.
However, both learning focuses exclusively on the occurrence of events (labels), while our framework involves a mixture distribution of $Y$ that simultaneously estimates occurrence and magnitude.

\vspace{1em}
\noindent{\bf Non-random missing labels:} There has been extensive work in learning with missing data. Our work is closely related to non-ignorable missing where the missing data mechanism depends on the unobserved values of a variable which is subject to missing \citep{Rubin1974-qx}. In this missing not at random (MNAR) setting, it is often required to specify a joint model for observations and missing mechanism in order to perform valid statistical inference. A number of works focus on model development, identifiability, estimation and predictions under various structural assumptions about missing mechanisms in the presence of MNAR outcomes \citep{Zhao2015-rb, Mohan2018-li, Franks2020-wd, Ma2019-ev}. Our work also concerns response-dependent missing labels, which are MNAR. However, true and false negatives are unknown in our setting, whereas which observations are missing is known a priori in the MNAR literature.

\vspace{1em}
\noindent{\bf Non-convex optimization and statistics:} Lastly, an active line of work exist in non-convex estimation problems in which various statistical and algorithmic guarantees of a non-convex M-estimator are studied \citep{Loh2012-tl, Yang2015-by, Mei2018-ec, Elsener2019-fo}. Our objective turns out to be a  non-convex function of parameters, and our work utilizes a number of tools in non-convex literature to obtain statistical and algorithmic guarantees of the proposed estimator which is a stationary point of the non-convex objective function.

\subsection{Contributions}
Our paper makes the following contributions:
\begin{enumerate}
\setlength{\itemsep}{0pt}%
\setlength{\parskip}{0pt}%
\setlength{\leftmargin}{0em}%
\item A general statistical framework for dealing with response-dependent missing labels, leading to a closed-form log-likelihood;
\item Identifiability conditions (Theorem~\ref{thm:1}) for our model;
\item We prove that any local minimizer achieves optimal (up to a constant) statistical error of $\frac{p}{n}$ (Theorem~\ref{thm:thm2}), where $p$ is the number of features and $n$ is the number of samples) under standard assumptions and proof that our projected gradient descent algorithm has geometric convergence to a local minimizer (Theorem~\ref{thm:thm3});
\item A simulation study which displays the advantages of our method compared to state-of-the-art methods under correct model specification and two misspecified model settings; and
\item Illustration of the advantages of our approach compared to existing stat-of-the-art approaches in a setting involving wildfire prediction in California.
\end{enumerate}

\subsection{Notation}
We use normal font for scalars (e.g. $a,b,c,\dots$) and boldface for vectors $(\x, \u, \v, \dots)$. We reserve capital letters for random variables. For a vector $\u, \v \in \R^d$, we write $\|\v\|_p$ to denote an $\ell_p$ norm of a vector. We also write $\B_q(r; \v)$ to denote an $\ell_q$ ball centered at $\v$, i.e. $\B_q(r;\v) := \{v; \|\v\|_q \leq r\}$. If the $\ell_q$ ball is centered at zero ($\v = 0$), we omit $\v$ and simply write $\B_q(r)$.

\section{Model and Algorithm}
\subsection{Problem Set-up}
We consider the following problem set-up for estimation and prediction using contaminated data. We assume that $Y$ has a mixture distribution of a point mass at $0$ (denoting no event) and continuous distribution over $\R_{+}$ (denoting the magnitude of the event), and each component distribution depends on the value of a set of features $\x \in \R^p$. In other words, the p.d.f of $Y$ given $X=\x$ is as follows\footnote{precisely by p.d.f, we mean a Radon-Nikodym derivative of $P_{Y|\X}$ with respect to the Lebesgue measure plus a point mass at zero.}:
\begin{align}\label{def:pdf_Y}
p_Y(t|\x; \beta,\theta ) = (1-p_1(\x))\delta_0(t) + p_1(\x) g(t|\x)
\end{align}
for some $p_1(\x)$ and $g(\cdot|\x)$ where $p_1$ takes a value between $0$ and $1$ depending on $\x$ and $g(t|\x)$ is a p.d.f of the continuous distribution. Here, each $p_1$ and $g$ is related to occurrence and magnitude of the mixture distribution for $Y$.

First, we model $\P(Y>0|\x) = p_1(\x;\theta)$ and $\P(Y=0|\x) = 1-p_1(\x;\theta)$ where we let $p_1(\x;\theta) := \sigma(\x^\top \theta):= (1+\exp(-\x^\top \theta))^{-1}$. When $Y>0$, we use an exponential GLM; specifically,
$$p_{Y|Y>0,\X}(y|\x) = g(t|\x;\beta) := \lambda_X \exp (-\lambda_X  t)$$ where $\lambda_X = \exp(-\x^\top \beta)$. The exponential GLM is chosen to reflect that a size of an event is always non-negative. That is, given that an event has occurred, i.e. $Y>0$, the probability that $Y$ is larger than $t$ is $\P(Y > t|Y>0, \x) = \int_t^\infty g(s|\x;\beta)ds$.

If an i.i.d sample of $(\x_i, y_i)_{i=1}^n$ is available, the mixture modeling approach (e.g. \citealp{Cragg1971-lh, Olsen2001-fz}) can be utilized to estimate the parameters $\theta$ and $\beta$. However in our setting, not all $y_i$ are observed since events with small magnitude that tend to have missing labels. We introduce a random variable $Z$ to denote the {\em observed} size of an event. If an event has occurred but is unobserved, then $y_i>0$ but $z_i = 0$.  On the other hand, if the event is observed, the recorded size is the same as the true size, i.e. $z_i = y_i$. Since $z_i = 0$ no longer implies that no event has occurred, we cannot simply estimate the parameters using the observed sizes ($z_i$s) instead of the true sizes ($y_i$s). 

\subsection{Likelihood model and identifiability}

We model the likelihood of correctly observing events as
\begin{align} \label{eq:gamma}
\P(Z>0|Y = y>0, \X=\x) =  \Gamma(y),
\end{align}
In other words, the probability that the magnitude $Y$ is observed depends only on the value of $Y$ itself.
In many practical applications, this ``self-masking phenomenon" occurs where true value itself determines whether the observation would be hidden or revealed. For example, if we consider fire prediction, the size of fire affects whether the fire event would be detected or not; hence $\Gamma(\cdot)$ is an monotonically increasing function.
From here, we  combine \eqref{def:pdf_Y} and \eqref{eq:gamma} and integrate out the unobserved $Y$ to derive $p_{Z|\X}$; the log of this quantity forms our loss function for a collection of samples $(\x_i,z_i)$ for $i=1,\ldots,n$:
\begin{align}\label{eq:logLik}
\begin{split}
&\L_n(\theta,\beta) = -\frac{1}{n}\sum_{i ; z_i = 0}
\log \left(1-\phi(\x_i; \beta, \Gamma)p_1(\x_i; \theta)\right)\\
&\qquad -\frac{1}{n}\sum_{i; z_i>0} \log \left\lbrace g(z_i|\x; \beta) \Gamma(z_i)p_1(\x_i;\theta)\right\rbrace
\end{split}
\end{align}
where
\begin{align}\label{def:phi}
\phi(\x; \beta, \Gamma) = \int_{0}^\infty \Gamma(y)g(y |\x; \beta) dy.
\end{align}

\noindent{\bf Identifiabiliy.} We first discuss the identifiability of the model. Clearly, the model is not identifiable if no assumptions about the structure of $g$ in \eqref{def:pdf_Y} and $\Gamma$ are made because the likelihood \eqref{eq:logLik} is defined via $\Gamma(y)g(y|\x)$. On the other hand, both parameters are identifiable under parametric assumptions on $p_1$ and $g$ for any given positive $\Gamma$, if two parameter vectors are distinct, i.e., $\beta\neq c\theta$ for any $c\neq 0$, and the feature vector $\x$ spans all directions in $\R^p$. More concretely, we state the following Assumption {\bfseries A\ref{a1}}:
\begin{assumption}\label{a1}
	Two parameter vectors $\beta$ and $\theta$ in \eqref{def:pdf_Y} are linearly independent. The density of $\P_X$ with respect to the Lebesgue measure is positive everywhere.
\end{assumption}
We have the following result about the identifiability of the model \eqref{eq:logLik}.
\begin{theorem}\label{thm:1}
For any given positive $\Gamma$ and under Assumption {\bfseries A\ref{a1}} , the parameters $(\beta,\theta)$ in the model \eqref{eq:logLik} are identifiable.
\end{theorem}
The proof is based on constructing a set of observations $(\x_i, z_i)$ which distinguish the likelihoods evaluated at different parameter values, and is deferred to the Supplementary Material.

\vspace{1em}
\noindent{\bf Choice of $\Gamma(\cdot)$.}  The next question is how to choose the label observation probability $\Gamma(y)$. One of the determining factors is that the integral in (\ref{def:phi}) needs to be computable and $\Gamma(y)$ also needs to be monotonically increasing. If $\phi$ does not have an analytical form, approximation of the function via a numerical integration is needed, which can be computationally challenging. Hence we choose $\Gamma$ to be the cumulative distribution function of an exponential function with parameter $\lambda_\epsilon$. In other words, we let
\begin{align}\label{def:Gamma}
\Gamma(y):= 1-\exp(-\lambda_\epsilon y).
\end{align}
We first note that $\Gamma$ is a monotonically increasing function in $y$. Therefore, events of larger magnitudes are more likely to be observed without noise. This choice of $\Gamma$ also allows a closed-form expression for $\phi(\cdot)$. More concretely, we have the following representation of $\phi$:
\begin{align}\label{def:phi2}
\begin{split}
\phi(\x; \beta, \Gamma) &= \int_{0}^\infty (1-e^{-\lambda_\epsilon y } ) e^{-\x^\top \beta} e^{-ye^{-\x^\top \beta}} dy \\
& = \frac{\lambda_\epsilon}{\lambda_\epsilon+e^{-\x^\top \beta}}.
\end{split}
\end{align}
Note that $\phi(\x; \beta, \Gamma) $ is a function of $\beta$ and $\lambda_\epsilon$ where the hyperparameter $\lambda_\epsilon$ controls the extent to which the labels $Y$ are missing.

Our estimation method is defined as the maximizer of the log-likelihood \eqref{eq:logLik} with $\Gamma$ and $\phi$ in \eqref{def:Gamma} and \eqref{def:phi2}. We use the name PU-OMM to refer to our method, which stands for {\em Positive-Unlabeled Occurrence Magnitude Mixture}.

\subsection{Algorithm}

Given data $(\x_i,z_i)$ for $i=1,\ldots,n$, the objective function is
\begin{align}\label{def:objective}
\widehat{\omega}  \in \argmin_{\omega \in \B_2(r)} \L_n(\omega):= -\frac{1}{n} \sum_{i=1}^{n}  \ell(\omega; (\x_i,z_i)),
\end{align}
where $\omega := (\beta,\theta)$, and $ \ell(\omega; (\x_i,z_i)) $ is the $i$th component of the likelihood in \eqref{eq:logLik} using the $\phi$ specified in \eqref{def:phi2}:
\begin{align*}
& \ell(\omega; (\x_i,z_i))
:= \\
&\quad \mathbbm{1}\{z_i= 0\}  \{\log \left(1-\phi(\x_i; \beta)p_1(\x_i; \theta)\right)\} +\mathbbm{1}\{ z_i>0\} \{\log  g(z_i|\x_i; \beta) + \log p_1(\x_i;\theta)\}.
\end{align*}
We also define the population risk function $\Rc(\omega) := \E[\L_n(\omega)]$ and define $\omega_0:= (\beta_0, \theta_0)$ as the minimizer of $\Rc(\omega)$. We let the search space $\B_2(r)$ be an $\ell_2$ ball with a radius $r$, for a sufficiently large $r>0$ so that $\omega_0$ is feasible.

To optimize \eqref{def:objective}, we propose to use the standard projected gradient descent (projected to $\B_2(r)$). We will show in Theorem \ref{thm:thm2} and \ref{thm:thm3} that it is feasible to obtain $\widehat{\omega}$ in \eqref{def:objective} despite  $\L_n(\omega)$ being non-convex, and the convergence of iterates $\{\omega^t\}_{t\geq 1}$  in Algorithm \ref{alg1-proj} is linear given a sufficiently large sample size.

 \begin{algorithm}[tbhp]
 	\SetAlgoLined
 	\KwIn{Data $(\x_i, z_i)_{i=1}^n$, step size $\eta$, initial point $\omega^0$, hyperparameter $\lambda_\epsilon$, search space radius $r$}
 	\For{$t = 1,2,3,\ldots$}{
 		$\omega^{t+1} = \mathcal{P}_{\B_2(r)} (\omega^{t} -\eta \triangledown \L_n(\omega^t))$\;
 		\If{converged}{STOP}
 	}
 	\caption{Projected Gradient Descent}\label{alg1-proj}
 \end{algorithm}

In practice, we fit the model using Algorithm \ref{alg1-proj} over a grid of $\lambda_\epsilon$ values. We chose the value of $\lambda_\epsilon$ which results in the best fit for the observed occurrence (See Implementation Details in Section 4 for more details). We empirically observed very good estimation and prediction performances of our model by choosing the hyperparameter in this way, where in many cases the performances of the models with the chosen $\lambda_\epsilon$s were comparable to the models with the true $\lambda_\epsilon$ values.

\section{Theoretical Guarantees}

Throughout this section, we assume that $\Gamma(t) = 1-\exp(-\lambda_\epsilon t)$ is given. We first introduce a set of conditions for the response variable, feature vector, and the degree of missingness, under which we prove algorithmic and statistical convergence.

\begin{assumption}\label{a2}
	(Random design) A random feature vector $\x \in \mathbb{R}^{p}$ with distribution $\P_X$ is mean-zero sub-Gaussian with parameter $K_X$ for a positive constant $K_X<\infty$. In other words, for any fixed unit vector $v \in \R^p$, we have
	\begin{align*}
	\E[ \exp (\x^\top v)^2 / K_X^2] \leq 2.
	\end{align*}
	Moreover, there exists $C_\lambda >0$ such that $\lambda_{\textnormal{min}}( \E[\x\x^\top]) \geq C_\lambda$.
\end{assumption}

\begin{assumption}\label{a3}
    (Boundedness) There exist constants $C_X, C_Y<\infty$ such that for the random feature $\x\in \R^p$ and response variable $y \sim p_Y(\cdot|\x;( \beta_0,\omega_0))$,
    $\|\x\|_2 \leq C_X$ and $|y/e^{\x^\top \beta_0}| \leq C_Y$ almost surely.
    \end{assumption}

Assumption 2 is a mild assumption on the feature vector $\x$ which states that $\x$ has a light probability tail and the smallest eigenvalue of the population covariance matrix is lower-bounded by a positive constant. The boundedness condition is assumed mainly for the technical convenience and states that both $\x$ and the deviation of $y$ from its mean are absolutely bounded, where we recall that $\E[Y|Y>0,\x] = e^{\x^\top \beta_0}$.

 \begin{assumption}\label{a4}
 	We assume the following condition holds:
 	\begin{align}\label{eq:a4}
 	\max_{1\leq i \leq n}  \frac{1 - \sigma( \x_i^\top \beta + \log \lambda_\epsilon)}{1-\sigma (\x_i^\top \theta)} \leq r_0(\omega_0, C_X, r)
 	\end{align}
where $r_0$ is a constant depending on model parameters $\omega_0 = (\beta_0, \theta_0)$, $C_X$, and $r$.
\end{assumption}

We give the full expression for $r_0(\omega_0, C_X, r)$ in the Supplementary Material for  ease of exposition. We recall that $\lambda_\epsilon = \infty$ corresponds to ``no missingness" where all $y_i$ are the same as $z_i$ since $\Gamma(y) = 1, \forall y$. The equation \eqref{eq:a4} trivially holds in this case. Assumption {\bfseries A\ref{a4}} essentially states that albeit $\lambda_\epsilon <\infty$, $\lambda_\epsilon$ is sufficiently large so that \eqref{eq:a4} holds. Assumption {\bfseries A\ref{a4}} ensures there exists sufficient signal in the data to estimate both parameters $\beta$ and $\theta$.

Under the stated assumptions, we first show that the population risk function $\Rc(\omega)$ has no other stationary point than $\omega_0$ in $\B_2(r)$ for a sufficiently large radius $r$ to include $\omega_0$.
\begin{proposition}\label{prop:prop2}
	Suppose Assumptions {\bfseries A\ref{a1}-A\ref{a4}} hold. Then for any $\omega \in \B_2(r)$ with $r \geq 2\|\omega_0\|_2$, we have,
	\begin{align}\label{eq:prop2}
	\langle \triangledown \Rc(\omega), \omega - \omega_0  \rangle  \geq \alpha \| \omega - \omega_0\|_2^2,
	\end{align}
	where the expectation is evaluated at the true parameter $\omega_0 = (\beta_0,\theta_0)$ and $\alpha>0$ is a positive constant depending only on the model parameters.
\end{proposition}

We defer the proof of this strong convexity result to the Supplementary Material. The essential step of the proof for  Proposition \ref{prop:prop2} is careful control of the size of a cross-product term which arises due to contamination in responses, to ensure a positive curvature of $\Rc(\omega)$ along $\omega-\omega_0$ directions.

Although the population risk function $\Rc(\omega)$ is non-convex, the inequality \eqref{eq:prop2} ensures that we can recover $\omega_0$ by finding a stationary point of $\Rc(\omega)$. Therefore, the population version of the algorithm \eqref{def:objective} is tractable. Together with the uniform convergence of the gradient $\triangledown \L_n(\omega)$, the equation \eqref{eq:prop2} immediately gives the bound for $\widehat{\omega}$ for any $\widehat{\omega}$ such that $\triangledown \L_n(\widehat{\omega}) = 0$ since
\begin{equation}\label{eq:8}
	\begin{aligned}
& \langle \triangledown \Rc(\widehat{\omega}) , \widehat{\omega} - \omega_0 \rangle \leq  \langle \triangledown\L_n(\widehat{\omega} ) - \triangledown \L_n( \omega_0), \widehat{\omega}-\omega_0\rangle \\
&\quad + 2\sup_{\omega \in \B_2(r)} \|\triangledown \L_n(\omega) - \triangledown\Rc(\omega)\|_2\|\widehat{\omega} - \omega_0\|_2
\end{aligned}
\end{equation}
and therefore,
\begin{align*}
\alpha \|\widehat{\omega} - \omega_0\|_2^2 & \leq \{\|\triangledown \L_n (\omega_0)\|_2  + 2a_n \}\|\widehat{\omega} - \omega_0\|_2,
\end{align*}
where $a_n :=\sup_{\omega \in \B_2(r)} \|\triangledown \L_n(\omega) - \triangledown\Rc(\omega)\|_2$. The rate of the statistical error bound $\|\widehat{\omega} - \omega_0\|$ is related to the order of (random) $\|\triangledown \L_n (\omega_0)\|_2$ and $a_n$. In particular, we can show that under Assumptions {\bfseries A\ref{a2} - A\ref{a4}}, both terms are of an order of $\sqrt{p \log (n)/n}$, therefore the bound we obtain is optimal up to a log term.

The following two theorems provide algorithmic and statistical error bounds.

\begin{theorem}\label{thm:thm2}
	 Under Assumptions {\bfseries A\ref{a1},-A\ref{a4}}, if $n \geq C p\log p$, the empirical risk function $\L_n(\omega)$ admits a unique local minimizer in $\B_2(r)$ which coincides with the global optimizer $\widehat{\omega}$. In addition, for any $\delta >0$, the following inequality holds with probability $1-\delta$,
\begin{align}\label{eq:thm2}
\|\widehat{\omega} - \omega_0\|_2 \leq \frac{C}{\alpha} \sqrt{\frac{C_Y^2 p \log (n) \log(C_Y/\delta)}{n}}
\end{align}
where $C>0$ is a constant only depending on model parameters and $\alpha$ is the constant in \eqref{eq:prop2} from Proposition \ref{prop:prop2}. 
\end{theorem}

\begin{theorem}\label{thm:thm3}
	 Assume {\bfseries A\ref{a1}-A\ref{a4}} hold. If $n \geq C p\log p$, for any initialization $\omega^0 \in \B_2(r/2)$,
	\begin{align}\label{eq:thm3}
	\|\omega^t - \widehat{\omega}\|_2 \leq C_1 \kappa^t \|\omega^0 - \widehat{\omega}\|_2
	\end{align}
	for $\kappa<1$, where $C, C_1>0$ are constants depending on model parameters (but not on $n,p$).
\end{theorem}

 The convergence rate in \eqref{eq:thm2} nearly matches the parametric rate of $\sqrt{p/n}$. Also, running the algorithm \ref{alg1-proj} efficiently finds the optimum of \eqref{def:objective}, in the sense that $O(\log(1/\epsilon))$ iterations are needed to find a point within  distance  $\epsilon$ of the global optimum $\widehat{\omega}$ of the objective \eqref{def:objective}.

\paragraph{Extension to the high-dimensional setting: }It is worth noting that the theory we develop here has a direct generalization to the high-dimensional setting where $p \gg n$  and we assume $\omega_0$ is $s$-sparse, for $s \ll p$. A similar approach as in \eqref{eq:8} can be used to obtain a statistical error bound of an $\ell_1$-penalized M-estimator $\widehat{\omega}(\lambda)$, defined as $\widehat{\omega}(\lambda):= \argmin \L_n(\omega) + \lambda \|\omega\|_1$, where we control the difference between $\L_n(\omega)$ and $\Rc(\omega)$ over a restricted cone including $\B_2(r)$  (see, for instance, \citealp{Mei2018-ec}) or equivalently we replace the strong convexity Proposition with restricted strong convexity \citep{Negahban2012-bd}.

\section{Simulation Study}
We now study the performance of the proposed method and compare with other state-of-the-art approaches in terms of parameter estimation accuracy and prediction using simulated datasets representing a number of scenarios. In particular, we consider the following three settings for generating simulated datasets where in the first setting our model is correctly specified and in the others different mis-specifications are introduced:
\begin{enumerate}
\setlength{\itemsep}{0pt}%
\setlength{\parskip}{0pt}%
\setlength{\leftmargin}{0em}%
	\item \textbf{Correct specification:} the size of an event $Y|(Y>0,\x)$ is generated from the exponential distribution with parameter $\lambda_X = \exp(-\x^\top \beta_0)$. Missing in $y_i$s are probabilistic, whose probabilities depend on $y_i$ via $\Gamma(y) = 1-\exp(-\lambda_\epsilon y)$ for $\lambda_\epsilon = .24$
	\item \textbf{Misspecification 1:} $g$ is log-Normal instead of exponential, i.e. $Y|(Y>0,\x) \sim \mbox{LogNormal}(\x_i^\top \beta_0,I_p)$.
	\item \textbf{Misspecification 2: } missing in $y_i$ is deterministic and $y_i$ below a certain threshold is recorded to be zero, i.e. $z_i = \mathbbm{1}\{y_i \geq \tau\}$ for a threshold $\tau=3$.
\end{enumerate}

\paragraph{Data Generation.}
We first generate a design matrix $\X = [\x_1^\top , \dots, \x_{n}^\top ]^\top $ by drawing each $\x_i$ from a multivariate Gaussian distribution $\mathcal{N}(0,\Sigma)$ where $\Sigma_{ij} = 0.2^{|i-j|}$. We sample two parameters from a normal distribution centered at zero, i.e. $\beta_0, \theta_0 \sim \mathcal{N}(0, 9p^{-1}I_p)$.
We sample the true unobserved response $y_i$ from a mixture of zero and a continuous distribution.

To do so, we first draw a binary $u_i \in\{0,1\}$ from a Bernoulli distribution whose probability depends on $\x_i^\top \theta_0$ to simulate the occurrence of an event, i.e.
$u_i \sim \mbox{Ber}(p_1(\x_i;\theta_0))$ where $p_1(\x_i; \theta_0) = (1+\exp(-\x_i^\top \theta_0))^{-1}$. If $u_i =1$, we draw the sample $v_i$ from a continuous distribution $g(\cdot|\x_i;\beta_0)$ for the size of an event. Depending on the setting, $g$ is set to be Exponential (Settings 1 and 3) or Lognormal (Setting 2). We let true responses $y_i$ to be $0$ if $u_i=0$ and $y_i = v_i$ otherwise.

Next, we sample $r_i \in \{0,1\}$ to determine whether each $y_i$ is missing or not. Depending on the setting, $\Gamma(y) = \P(R = 1|y) = 1-e^{-\lambda_\epsilon y}$ for $\lambda_\epsilon = .24$ (Settings 1 and 2) or $\Gamma(y) =  \mathbbm{1}\{y_i \geq \tau\}$ for $\tau = 3$ (Setting 3). The observed response $z_i$ is set to be $z_i = y_i r_i$ so that if $r_i = 1$, then $z_i = y_i$ and if $r_i= 0$, then $z_i=0$.

For each simulation trial $b=1,\dots,B=50$, the final datasets include $(\x_i,z_i)_{i=1}^n$ for the training dataset and $(\x_i,z_i)_{i=1}^{n_{\rm test}}$ for the test dataset. We additionally keep $y_i$ values in the training and test datasets for the purpose of fitting an oracle model (with no missing in $y_i$) for comparison, and validate prediction performances for the \textit{true responses}. We let $p = 10$ and vary $n$ from $5000$ to $30000$ for a training set, and let $n_{\rm test}=50000$ for a test set for the accurate evaluation of trained models.

\paragraph{Methods.}

\begin{enumerate}
\setlength{\itemsep}{0pt}%
\setlength{\parskip}{0pt}%
\setlength{\leftmargin}{0em}%
	\item Oracle: two GLMs (Logistic, Exponential) using $(\x_i, y_i)_{i=1}^n$ where $y_i$ with fully labelled responses.
	\item Proposed method (PU-OMM): our proposed method.
	\item Logistic-Gamma mixture model (Logistic-Gamma): we fit two separate GLMs using $(\x_i, z_i)_{i=1}^n$, one for the occurrence and the other for the size of the event using logistic and Gamma distributions
	\item Logistic-LogNormal mixture model (Logistic-LogNormal): Gamma distribution is replaced with log-normal distribution in 3.
\end{enumerate}

\paragraph{Evaluation Metrics.} We evaluate both parameter estimation accuracy for $\beta_0,\theta_0$ and prediction accuracy for estimated occurrence and size of true events.
For parameter estimation accuracy, We compute Root Mean Squared Errors (RMSE) for each estimated $(\widehat{\beta}, \widehat{\theta})$:
\begin{itemize}
	\item RMSE(beta):= $\|\widehat{\beta} - \beta_0\|_2$
	\item RMSE(theta):= $\|\widehat{\theta} - \theta_0\|_2$
\end{itemize}

We also evaluate the prediction accuracy of each model in terms of predicting both occurrence and size of the \textit{true} events. For predicting occurrence of an event, we use the following two metrics:
\begin{itemize}
	\item BrierLoss$(\u,\widehat{\mathbf{p}})$:= $\frac{1}{n_{\rm test}} \sum_{i=1}^{n_{\rm test}} (\widehat{p}_i- u_i )^2$
	\item Misclassification$(\u, \widehat{\mathbf{p}})$:= $\frac{1}{n_{\rm test}} \sum_{i=1}^{n_{\rm test}} \mathbbm{1}\{\mathbbm{1} \{\widehat{p}_i>0.5\} \neq u_i\}$
\end{itemize}
where $\u \in \R^{n_{\rm test}}$ is a vector of indicator variables where each $u_i:=\mathbbm{1}\{y_i>0\}$ represents the occurrence of an event, and $\widehat{p}_i$ is a predicted probability for the occurrence of $i$th observation from each model.

For predicting the magnitude of an event, we use Mean Absolute Deviation (MAD), root mean squared error (RMSE), and Symmetric Mean Absolute Percentage Error (SMAPE) for prediction evaluation metrics. SMAPE is considered to evaluate prediction performance also in a relative scale, as results of MAD and RMSE can be affected by a few observations with large errors \citep{Chen2017-oq}.
MAD, RMSE, and SMAPE between realized values $\mathbf{y}\in\R^{n_{\rm test}}$ and predicted values $\widehat{\mathbf{y}} \in \R^{n_{\rm test}}$ are computed as follows:
\begin{itemize}
	\item MAD($\mathbf{y}, \widehat{\mathbf{y}}$):= $\frac{1}{n_{\rm test}} \sum_{i=1}^{n_{\rm test}} |y_i - \widehat{y}_i|$
	\item RMSE($\mathbf{y}, \widehat{\mathbf{y}}$):= $\sqrt{ \frac{1}{n_{\rm test}} \sum_{i=1}^{n_{\rm test}} (y_i - \widehat{y_i} )^2}$
	\item SMAPE($\mathbf{y}, \widehat{\mathbf{y}}$):= $\frac{1}{n_{\rm test}} \sum_{i=1}^{n_{\rm test}} \frac{2|y_i - \widehat{y}_i|}{|y_i|+|\widehat{y}_i|}$
\end{itemize}
where $\hat{y}_i$ is the predicted value from each model for the size of an event of the $i$th observation in a test dataset.  
\begin{figure*}[htb]
	\centering
	\begin{subfigure}{1\textwidth}
		\centering
			\includegraphics[width=.95\linewidth]{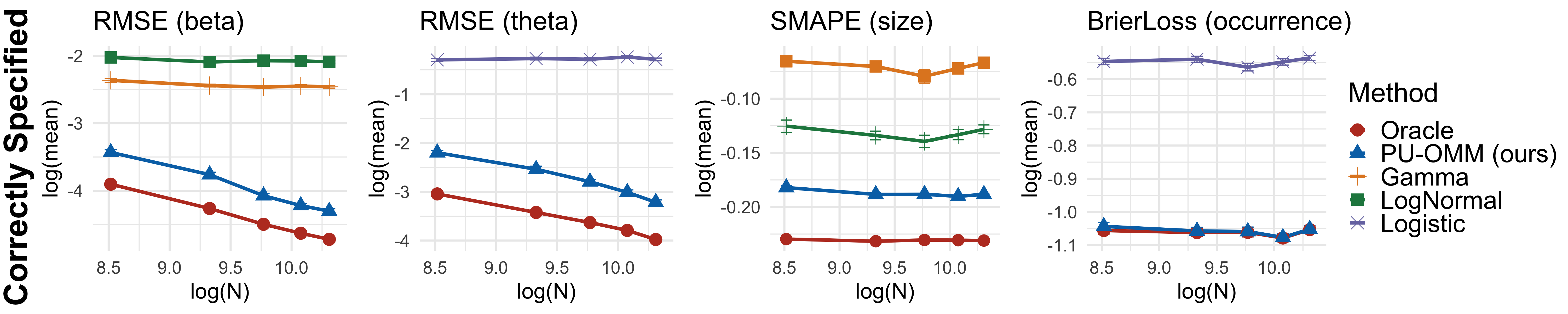}
		\label{fig:setting1}
	\end{subfigure}
\begin{subfigure}{1\textwidth}
	\centering
	\includegraphics[width=.95\linewidth]{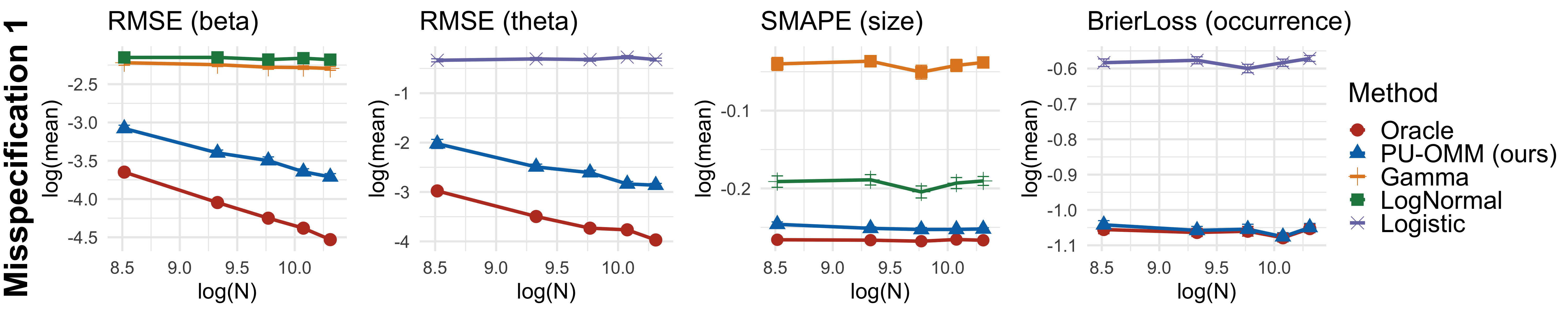}
	\label{fig:setting2}
\end{subfigure}
	\begin{subfigure}{1\textwidth}
		\centering
			\includegraphics[width=.95\linewidth]{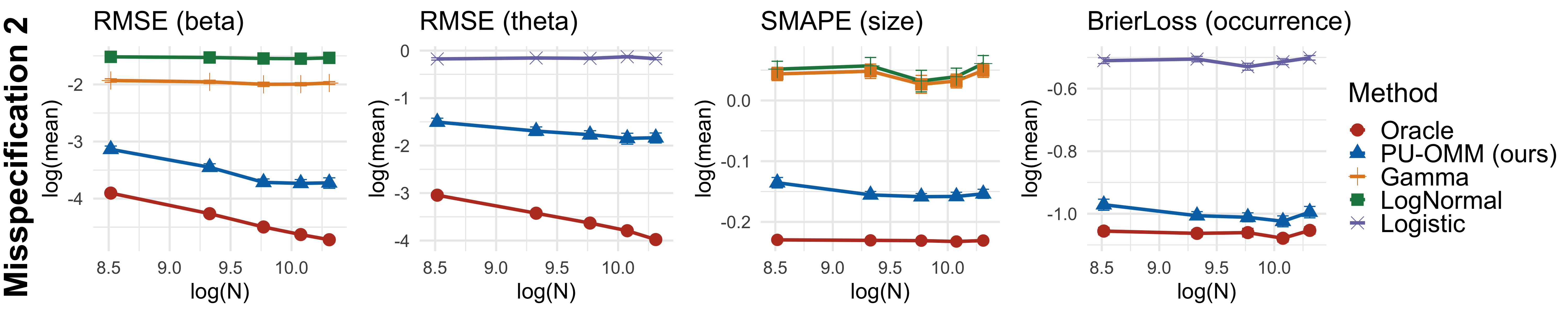}
		\label{fig:setting3}
	\end{subfigure}
	\caption{Parametric estimation and prediction accuracy for each method under Settings 1-3 (Correctly Specified, Misspecification 1, and Misspecification 2). Each row $i$ corresponds to the Setting $i$, for $i=1,2,3$. For each row, first two panels (RMSE (beta), RMSE (theta)) show  parameter estimation accuracy results, and the last two panels (SMAPE (size), BrierLoss (occurrence)) plot the  accuracy of each method in predicting the true size and occurrence of each observation in test datasets. Average values from $B=50$ trials are plotted, together with error bars corresponding to one standard error. Note for the Logistic-Gamma and Logistic-LogNormal Mixture models, the logistic model is used to predict the occurrence of events, and Gamma/LogNormal model is used to predict the magnitudes of events. Therefore, our PU-OMM model is compared with the Gamma/LogNormal models in RMSE (beta) and SMAPE (size) panels, and PU-OMM is compared with the Logistic model in RMSE (theta) and BrierLoss (occurrence) panels.}\label{fig1:evalModels}
\end{figure*} 

\paragraph{Implementation Details.}
PU-OMM is fitted over a grid of 20 $\lambda_\epsilon$ values from $1/50$ to $50$ which are linearly spaced on a log scale. For each value of $\lambda_\epsilon$, the objective \eqref{def:objective} for the proposed method is minimized via the projected gradient descent method in Algorithm \ref{alg1-proj}, where we let $r = 5\sqrt{p}$ for the radius of the search region. A backtracking-Armijo linesearch is performed at each iteration to ensure a sufficient decrease \citep{Beck2017-us}.

Once we have $20$ fitted models, we chose the best $\lambda_\epsilon$ based on goodness of fit for the observed occurrence. More concretely, we chose $\lambda_\epsilon$ at the value where the fitted model minimizes the Brier Loss for the observed occurrences in the training dataset, i.e.
\begin{align*}
	\widehat{\lambda}_\epsilon := \argmin_{\lambda  \in \{ 1/50,\dots,50\} } {\rm Brier Loss} (\v, \widehat{\mathbf{q}}(\lambda ))
\end{align*}
where $\v \in \{0,1\}^{n}$ is a vector of the observed occurrences in the training dataset, i.e. $v_i := \mathbbm{1}\{z_i>0\}$, and $\widehat{q}(\lambda)_i$ is the predicted probability for the observed occurrence $\mathbbm{1}\{z_i>0\}$ with $\lambda_\epsilon = \lambda$.

\paragraph{Results.}
Figure \ref{fig1:evalModels} presents estimation and prediction accuracy for each method under Settings 1-3. We plot results using SMAPE and BrierLoss in Figure \ref{fig1:evalModels} for prediction evaluation and defer the remaining plots to the Supplementary Material. Unsurprisingly, the oracle estimator performs the best. Among non-oracle methods, the proposed method appears to perform the best in both correctly specified and misspecified settings, even when the hyperparameter $\lambda_\epsilon$ is chosen based on the data. In fact, the difference between the two PU-OMM models--one based on the true $\lambda_\epsilon$ value and the other based on the choice from data--was quite small. We also include a comparison plot between the two PU-OMM models in the Supplementary Material.

\section{California Wildfire Data}
\subsection{California Wildfire Dataset}
We use a global wildfire dataset from \cite{Artes2019-ay} to obtain observed fire events in California from 2001 to 2018. The database \cite{Artes2019-ay} includes fire events--sets of burnt areas that are connected by touching or intersecting--together with fire perimeters and the final dates of the fire events. We obtain fire sizes by computing areas of fire events based on fire perimeters.
 
75\% of the observed fires have sizes ranging from .19km$^2$ to $1.33$km$^2$, whereas the smallest was $.0003$ km$^2$ and the largest was $1083.82$ km$^2$. More importantly, most of the fires whose sizes are below $1$km$^2$ are not present in the dataset. Since we expect there would be more small fires than large fires, it is likely that fires smaller than .19km$^2$ are not recorded due to the limitations in data collection and processing accuracy.

Given the lack of small fires in the database, we additionally sampled points from places with no observed fires. To be more specific, for each year, we uniformly sampled the same number of points as the observed fires from the map of California excluding .1 degree buffered fire events. We augmented the fire events dataset from \cite{Artes2019-ay} by adding these points where the fire sizes corresponding to these points are set to be zero.

We also incorporated information on meteorological, topographical, geographical aspects of each sampled location. Specifically, we included elevation, slope, aspect, dissection, heat load index, topographic position index, and terrain ruggedness index from the STRM 90m resolution data \citep{Jarvis2008-yp} for topography-related variables, daily temperature, precipitation, relative humidity, and vapour-pressure deficit (VPD) from ERA-interim reanalysis data \footnote{\sloppypar available at https://www.ecmwf.int/en/forecasts/datasets /reanalysis-datasets/era-interim}, and population density, distance to the closest high population density area, and distance to the closest low population density area from Gridded Population of the World (GPW) \citep{CIESIN-2017}. The final dataset has dimensions $(n,p) =  (15846,43)$.

\subsection{Results}
\begin{figure}[tbp]
	\includegraphics[width=1\linewidth]{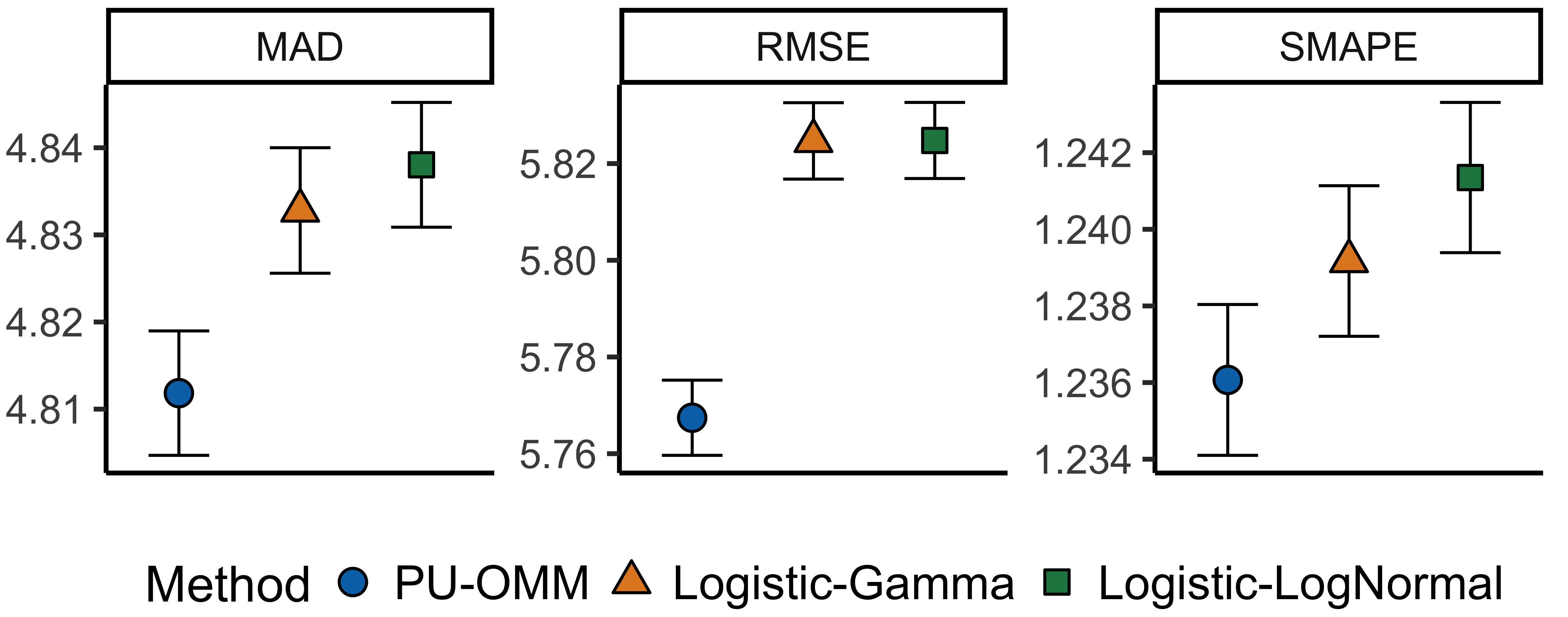}
	\caption{Prediction performance comparison for PU-OMM, Logistic-Gamma, and Logistic-LogNormal models with the California Wildfire dataset. Average MAD, RMSE, and SMAPE values are plotted for each method. Error bars represent 1 standard error.}
	\label{fig:calfirecomparison}
\end{figure}

\paragraph{Prediction Accuracy}
All of the models are trained based on a training dataset and tested on the remaining hold-out set. For each $b =1,\dots,B = 100$, we randomly split the dataset into 90/10 subsamples and assigned 90\% of the subsamples to a training dataset and the remaining 10\% of the subsamples to a testing dataset.

Unlike the simulated study, true $y_i$ are unavailable, and thus validation needs to be  based on the observed $z_i$. We compute predicted $\widehat{z}_i$ using fitted models. In particular, MAD($\mathbf{z}, \widehat{\mathbf{z}}$), RMSE($\mathbf{z}, \widehat{\mathbf{z}}$), and SMAPE($\mathbf{z}, \widehat{\mathbf{z}}$) are computed based on the observed $z_i$ and predicted $\widehat{z}_i$. Figure \ref{fig:calfirecomparison} plots computed MAD, RMSE, and SMAPE from various models from $B$ trials. It appears that the proposed PU-OMM method performs the best, followed by Logistic-Gamma, and then followed by Logistic-LogNormal model.

\section{Discussion and Conclusion}
In this paper, we developed a general statistical framework PU-OMM which addresses occurrence and magnitude prediction when we have response-dependent missing labels. We prove that our approach achieves optimal statistical error up to a log factor, even though the likelihood loss is non-convex. Moreover, we also showed that our projected gradient descent algorithm achieves linear convergence to a stationary point of the objective. Also as discussed in Section 3, our framework and statistical and algorithmic guarantees have direct generalization to the high-dimensional setting.

Our flexible framework can be generalized to other response-dependent missing labels settings where the missing mechanism is a stochastic function of the response values but with different models of the occurrence-magnitude mixture response. This extra flexibility comes with statistical and algorithmic challenges such as computing the integral required for the log-likelihood and providing guarantees for the non-convex objective. Adapting this framework to other missing labels settings remains an open challenge.


\bibliographystyle{plainnat}
\bibliography{reference}
\newpage
\appendix
\begin{center}
  {\bfseries SUPPLEMENTARY MATERIAL}
\end{center}
\section{Proofs}
\subsection{Proof of Theorem 1}
For any given $\Gamma>0$, we show that $g(t|\x;\beta)\Gamma(t) p_1(\x;\theta) = g(t|\x;\tilde{\beta}) \Gamma(t)p_1(\x;\tilde{\theta})$ for all $t \in \R$ and $\x \in \R^p $ implies $\beta = \tilde{\beta}$ and $\theta = \tilde{\theta}$.
\begin{align*}
&g(t|\x;\beta)\Gamma(t) p_1(\x;\theta) = g(t|\x;\tilde{\beta}) \Gamma(t)p_1(\x;\tilde{\theta}) \\
&\Leftrightarrow  \log g(t|\x;\beta)+ \log p_1(\x;\theta) = \log g(t|\x;\tilde{\beta})+  \log p_1(\x;\tilde{\theta}) ,
\end{align*}
for all $t>0$ and $\x \in \R^p$.

Note that $ \log g(t|\x;\beta)$ and $\log p_1(\x;\theta)$ is a function of $\x^\top \beta$ and $\x^\top \theta$, since $\log g(t|\x;\beta) = -\x^\top \beta -\exp(-\x^\top\beta )t$ and $\log p_1(\x;\theta) = \x^\top \theta - \log (1+\exp(\x^\top \theta))$. For any $t>0$, we have,
\begin{align}\label{eq:pthm1-1}
-\x^\top \beta - \exp(-\x^\top \beta )t + \x^\top \theta   - \log (1+\exp(\x^\top \theta))= -\x^\top \tilde{\beta} - \exp(-\x^\top \tilde{\beta} )t + \x^\top\tilde{\theta} - \log (1+\exp(\x^\top \tilde{\theta}))
\end{align}

From \eqref{eq:pthm1-1}, we want to conclude that $\beta = \tilde{\beta}$ and $\theta = \tilde{\theta}$.
First, we let $t =0$ to obtain,
\begin{align*}
\x^\top(\theta-\beta) - \log(1+\exp(\x^\top\theta)) = \x^\top(\tilde{\theta}-\tilde{\beta}) - \log(1+\exp(\x^\top\tilde{\theta}))
\end{align*}
Let $\alpha_1 = \theta-\beta, \alpha_2  =\theta$. Obtain $\{\alpha_j\}_{3\leq j \leq p}$ such that $\mathbf{A}:= [\alpha_1,\alpha_2,\dots,\alpha_p] \in \R^{p \times p}$ be an invertible matrix. Define $\delta_1 = \mathbf{A}^{-1}(\tilde{\theta}-\tilde{\beta})$, $\delta_2 = \mathbf{A}^{-1}\tilde{\theta}$. For any $\x$ and $\mathbf{u} := \mathbf{A}^\top \x  = [u_1,\dots,u_p]^\top \in \R^p$,
\begin{align*}
u_1 - \log(1+\exp(u_2)) &= \mathbf{u}^\top \mathbf{A}^{-1}(\tilde{\theta}-\tilde{\beta}) - \log(1+\exp(\mathbf{u}^\top \mathbf{A}^{-1}\tilde{\theta}))\\
&= \mathbf{u}^\top \delta_1 - \log(1+\exp(\mathbf{u}^\top \delta_2)).
\end{align*}
We choose $\{\x^{(j)}\}_{1 \leq j \leq p}$ so that for each $\x^{(j)}$, $u_j^{(j)} =\mathbf{A}^\top x_{j}^{(j)} \neq 0, u_k^{(j)} = \mathbf{A}^\top x_{k}^{(j)} = 0, \forall k \neq j$.
For $j=1$ and any $s\neq 0$, we have,
\begin{align}\label{eq:1_1}
su_1^{(1)} - \log(2) &= su_1^{(1)} \delta_{11} - \log(1+\exp(su_1^{(1)} \delta_{21}))
\end{align}
Viewing \eqref{eq:1_1} as a function of $s$, we conclude that $\delta_{11}=1, \delta_{21} = 0$. Similarly, for $j = 2$,
\begin{align*}
- \log(1+\exp(su_2^{(2)})) &= su_2^{(2)} \delta_{12} - \log(1+\exp(su_2^{(2)} \delta_{22}))
\end{align*}
and obtain $\delta_{12}=0, \delta_{22}=1$. For $j\geq 3$, we have,
\begin{align*}
- \log(2) &= su_2^{(j)} \delta_{1j} - \log(1+\exp(su_2^{(j)} \delta_{2j})),
\end{align*}
thus $\delta_{1j} = \delta_{2j}= 0, \forall j\geq 3$. In other words, $\delta_1 = e_1, \delta_2 = e_2$ where $e_j$ is the $j$th canonical basis vector.
From the definition of $\delta_1$ and $\delta_2$, we have, $ \tilde{\theta}-\tilde{\beta}=  \mathbf{A}\delta_1 = \mathbf{A}e_1 = \theta-\beta$ and $ \tilde{\theta} = \mathbf{A}\delta_2 =  \mathbf{A}e_2 = \theta $. Therefore we conclude that $ \theta = \tilde{\theta}$, $\beta = \tilde{\beta}$.

\subsection{Derivation of the Likelihood}
Ler $R \in \{0,1\}$ be a binary random variable such that $R|(\X=x,Y=y) \sim Ber(\Gamma(y))$ for a continuous $\Gamma$ supported on $[0,\infty]$ and $\Gamma(0)=0$. On $R=1$, $Z=Y$ a.s., and on $R=0$, we let $Z = 0$.

We first compute the cdf of $Z$. First, for $t <0$, $\P(Z\leq t | \X=\x) = 0$.  For any $t\geq 0$,
\begin{align*}
\P(Z\leq t | \X=\x)
&=  \P(Z \leq t, R = 0|\X = \x)+\P(Z  \leq t, R = 1 |\X = \x) \\
& = \P(Z  \leq t , R = 0|\X = \x)+\P(Y \leq t, R = 1|\X = \x)
\end{align*}
We first address the second term. We have,
\begin{align*}
\P(Y \leq t, R = 1|\X = \x)  &=   \int_0^t \P(R=1|\X=\x,Y=y) p_Y (y|\x) d(m+\delta_0)\\
& =\P(R=1|\X=\x,Y=0) (1-p_1(\x;\theta)) \\
&\quad+\int_0^t\P(R=1|\X=x,Y=y) p_1(\x;\theta) g(y|\x;\beta) dy
\end{align*}
where $m + \delta_0$ is the Lebesgue measure \emph{plus } a point mass at zero. We have $\P(R=1|\X=x,Y=0) =\Gamma(0)= 0$ and $\P(R=1|\X=x,Y=y) =\Gamma(y)$ for $y>0$. Therefore,
\begin{align}\label{eq:lik1}
\P(Z \leq t, R = 1|\X = \x)  = p_1(\x;\theta) \int_0^t \Gamma(y)  g(y|\x;\beta) dy
\end{align}
Now we address the first term:
\begin{align*}
\P(Z  \leq t , R = 0|\X = \x) =  \P(Z = 0, R = 0|\X = \x) =  \P(R = 0|\X = \x).
\end{align*}
Since,
\begin{align*}
\P(R = 0|\X = x)
&= \int_{(0,\infty]} \P(R=0|\X=\x,Y=y) p_Y (y|\x) d(m+\delta_0) \\
&= \P(R = 0|Y = 0, \X = x)(1-p_1(\x;\theta)) \\
&\quad+ \int_{(0,\infty]} \P(R = 0|Y = y, \X = x)p_1(\x;\theta)g(y|\x;\beta)dy\\
&= 1-p_1(\x;\theta)+\int_{(0,\infty]} \{1-\Gamma(y)\}p_1(\x;\theta)g(y|\x;\beta)dy\\
& = \{1-p_1(\x;\theta) \}+p_1(\x;\theta) - p_1(\x;\theta) \int_{(0,\infty]} \Gamma( y)g(y|\x;\beta)dy,
\end{align*}
we have,
\begin{align}\label{eq:lik2}
\P(Z \leq t | \X =x) = \{1-p_1(\x;\theta) \}+p_1(\x;\theta) - p_1(\x;\theta) \int_{(0,\infty]} \Gamma( y)g(y|\x;\beta)dy.
\end{align}

Combining \eqref{eq:lik1} and \eqref{eq:lik2}, for any $t\geq 0$,
\begin{align*}
\P(Z \leq t | \X =x) &=  1-  p_1(\x;\theta) \int_{(0,\infty]} \Gamma( y)g(y|\x;\beta)dy + p_1(\x;\theta) \int_0^t \Gamma(y)  g(y|\x;\beta) dy\\
\end{align*}
and therefore the conditional pdf of $Z$ (with respect to $m+\delta_0$) is
\begin{align*}
f_Z(t|\x) = \delta_0(t)\lbrace 1- p_1(\x;\theta)\int_{(0,\infty]} \Gamma(y)g(y|\x;\beta)dy\rbrace+ \mathbbm{1}\{t>0\} g(t|\x;\beta) \Gamma(t) p_1(\x;\theta).
\end{align*}

\subsection{Proof of Proposition 1}
Let $\omega= (\beta,\theta) \in \B_2^{2p}(r)$ be given. We let $U_i := \1{Z_i >0}$ and $u_i$ be a realization of $U_i$. From the equation below (7) in the main text,
\begin{align*}
\ell(\omega; (\x_i,z_i))&=  (1-u_i)
\log \left(1-\sigma(\x_i^\top \beta+\log \lambda_\epsilon)\sigma(\x_i^\top \theta)\right)+ u_i \left\lbrace -\exp(-\x_i^\top\beta) z_i -\x_i^\top \beta +	\log \sigma(\x_i^\top \theta)) \right\rbrace\\
& = u_i \log \sigma(\x_i^\top \theta) + (1-u_i) \log \left(1-\sigma(\x_i^\top \beta+\log \lambda_\epsilon)\sigma(\x_i^\top \theta)\right)-u_i \{e^{-\x_i^\top\beta} z_i +\x_i^\top \beta \},
\end{align*}
since
\begin{align*}
\phi(\x;\beta, \Gamma) &=\int_{0}^\infty (1-e^{-\lambda_\epsilon y } ) \lambda_X e^{-y \lambda_X} dy = 1-\frac{\lambda_X}{\lambda_X+\lambda_\epsilon} =  \frac{\lambda_\epsilon e^{\x^\top\beta}}{\lambda_\epsilon e^{\x^\top\beta}+1} = \sigma(\x^\top \beta + \log \lambda_\epsilon),
\end{align*}
where $\lambda_X = e^{-\x^\top \beta}$.

We add and subtract $u_i \log \sigma(\x_i^\top \beta + \log \lambda_\epsilon)$ to have,
\begin{align*}
\ell(\omega; (\x_i,z_i))&= u_i \log \sigma(\x_i^\top \theta) \sigma(\x_i^\top \beta + \log \lambda_\epsilon) + (1-u_i) \log \left(1-\sigma(\x_i^\top \beta+ \log \lambda_\epsilon)\sigma(\x_i^\top \theta)\right)\\
&-u_i (e^{-\x_i^\top\beta} z_i +\x_i^\top \beta +\log \sigma(\x_i^\top \beta+ \log \lambda_\epsilon)).
\end{align*}
Letting $p_i = \sigma(\x_i^\top \theta) \sigma(\x_i^\top \beta+\log \lambda_\epsilon)$, we note that the first two terms have the form
\begin{align*}
u_i \log (p_i)  + (1-u_i ) \log (1-p_i) = u_i \log \frac{p_i}{1-p_i} + \log(1-p_i).
\end{align*}
Define
\begin{align}\label{def:h}
h(x,y) :=\log \left(  \frac{\sigma(x)\sigma(y)}{1-\sigma(x)\sigma(y)} \right)
\end{align}
so that
\begin{align}
 \log \frac{p_i}{1-p_i} = h(\x_i^\top\beta+\log\lambda_\epsilon, \x_i^\top \theta) =  \log\frac{ \sigma(\x_i^\top \theta) \sigma(\x_i^\top \beta+\log \lambda_\epsilon)}{1- \sigma(\x_i^\top \theta) \sigma(\x_i^\top \beta+\log \lambda_\epsilon)}. \nonumber
\end{align}
 
For the ease of notation, we let $f_i(\beta,\theta) := f(\x_i^\top\beta + \log \lambda_\epsilon , \x_i^\top \theta)$ for any function $f: \R\times\R \rightarrow \R$. For instance, we let
$h_i(\beta,\theta)
= h(\x_i^\top\beta+\log\lambda_\epsilon, \x_i^\top \theta)$. We have,
\begin{align*}
\ell(\omega; (\x_i,z_i))= u_i h_i(\beta,\theta) - \log (1+ e^{h_i(\beta,\theta)})-u_i \{e^{-\x_i^\top\beta} z_i +\x_i^\top \beta+\log \sigma(\x_i^\top \beta+\log \lambda_\epsilon)\}.
\end{align*}
Now we take derivatives with respect to $\beta$ and $\theta$.
\begin{align*}
\frac{\partial}{\partial\beta} 	\ell(\omega; (\x_i,z_i))&= \left \lbrace u_i - \frac{e^{h_i(\beta,\theta)}}{1+e^{h_i(\beta,\theta)}}\right\rbrace h_{1i}(\beta,\theta)\x_i - u_i \left\lbrace  - e^{-\x_i^\top \beta} z_i +1+\frac{1}{1+e^{\x_i^\top \beta+\log \lambda_\epsilon}}\right\rbrace \x_i\\
\frac{\partial}{\partial\theta}	\ell(\omega; (\x_i,z_i))&= \left \lbrace u_i - \frac{e^{h_i(\beta,\theta)}}{1+e^{h_i(\beta,\theta)}}\right\rbrace h_{2i}(\beta,\theta)\x_i.
\end{align*}
where $h_1$ and $h_2$ are partial derivatives of $h$ with respect to the first and second arguments, i.e.
\begin{align}\label{eq:h1_h2}
h_1(x,y) = \frac{1-\sigma(x)}{1-\sigma(x)\sigma(y)} \quad \mbox{, and} \quad	h_2(x,y) = \frac{1-\sigma(y)}{1-\sigma(x)\sigma(y)}.
\end{align}

Then,
\begin{align*}
\langle - \triangledown \Rc(\omega) , \omega - \omega_0 \rangle
&= \langle   \triangledown \E[\ell(\omega; (\x_i,z_i))],\omega - \omega_0\rangle \\
&= \left \langle \begin{bmatrix}
\E[\frac{\partial}{\partial\beta}\ell(\omega; (\x_i,z_i))] \\
\E[\frac{\partial}{\partial\theta}\ell(\omega; (\x_i,z_i))]\\
\end{bmatrix}
,
\begin{bmatrix}
\beta-\beta_0\\
\theta - \theta_0\\
\end{bmatrix}
\right \rangle\\
& = \E[\frac{\partial}{\partial\beta} \ell(\omega; (\x_i,z_i))^\top(\beta-\beta_0)]+\E[\frac{\partial}{\partial\theta} \ell(\omega; (\x_i,z_i))^\top(\theta-\theta_0)].
\end{align*}
where we exchange the derivative and expectation, which is valid by dominated convergence theorem and Assumption {\bfseries A3}.
First, we let $\Delta_\theta = \theta - \theta_0 $ and $\Delta_\beta = \beta - \beta_0$. By the law of iterated expectations,
\begin{align*}
\E[\frac{\partial}{\partial\theta} \ell(\omega; (\x_i,z_i))^\top(\theta-\theta_0)]
&= \E\left[\left \lbrace u_i - \frac{e^{h_i(\beta,\theta)}}{1+e^{h_i(\beta,\theta)}}\right\rbrace h_{2i}(\beta,\theta)\x_i^\top (\theta-\theta_0) \right]\\
&= \E\left[\left \lbrace \E[u_i|\x_i] - \frac{e^{h_i(\beta,\theta)}}{1+e^{h_i(\beta,\theta)}}\right\rbrace h_{2i}(\beta,\theta)\x_i^\top \Delta_\theta \right],
\end{align*}
and
\begin{align*}
\E[\frac{\partial}{\partial\beta} \ell(\omega; (\x_i,z_i))^\top(\beta-\beta_0)] &= \E\left[\left \lbrace \E[u_i|\x_i] - \frac{e^{h_i(\beta,\theta)}}{1+e^{h_i(\beta,\theta)}}\right\rbrace h_{1i}(\beta,\theta)\x_i^\top \Delta_\beta \right.\\
&  \quad + \left. ( \E[ z_i u_i|\x_i] e^{-\x_i^\top \beta} - \E[u_i|\x_i] (1+\frac{1}{1+e^{\x_i^\top \beta+\log \lambda_\epsilon}}) ) \x_i^\top \Delta_\beta \right].
\end{align*}
From the definition of $h$ in \eqref{def:h}, we have,
\begin{align*}
\E[u_i|\x_i] = \frac{e^{h(\x_i^\top\beta_0, \x_i^\top \theta_0)}}{1+e^{h(\x_i^\top\beta_0, \x_i^\top \theta_0)}} = \sigma(\x_i^\top \theta_0) \sigma(\x_i^\top \beta_0+\log \lambda_\epsilon),
\end{align*}
and thus,
\begin{equation}\label{eq:direct_deriv}
\begin{aligned}
\E[\frac{\partial}{\partial\omega} \ell(\omega; (\x_i,z_i))^\top(\omega-\omega_0)]
&=
\E\left[\left \lbrace A'({h(\x_i^\top\beta_0, \x_i^\top \theta_0)}) - A'(h_i(\beta,\theta)) \right\rbrace h_{1i}(\beta,\theta)\x_i^\top \Delta_\beta \right]\\
&+\E\left[\left \lbrace A'({h(\x_i^\top\beta_0, \x_i^\top \theta_0)}) - A'(h_i(\beta,\theta)) \right\rbrace h_{2i}(\beta,\theta)\x_i^\top \Delta_\theta \right]	\\
&+\E \left[(\E[ z_i u_i|\x_i] e^{-\x_i^\top \beta} - \E[u_i|\x_i] (1+\frac{1}{1+e^{\x_i^\top \beta+\log \lambda_\epsilon}}) ) \x_i^\top \Delta_\beta\right],
\end{aligned}
\end{equation}
where we define $A(t) = \log (1+\exp(t))$.

For the sum of the first two terms (:= Term I) in \eqref{eq:direct_deriv},
\begin{align*}
\textnormal{Term I} := &\E\left[\left \lbrace A'({h(\x_i^\top\beta_0, \x_i^\top \theta_0)}) - A'(h(\x_i^\top\beta,\x_i^\top\theta)) \right\rbrace
\begin{bmatrix}
h_{1i}(\beta,\theta)\\
h_{2i}(\beta,\theta)
\end{bmatrix}^\top
\begin{bmatrix}
\x_i^\top \Delta_\beta \\
\x_i^\top \Delta_\theta
\end{bmatrix}
\right] \\
& = - \E \left[A''(h(\x_i^\top\beta_i, \x_i^\top \theta_i))
\left(
\begin{bmatrix}
h_{1i}(\beta_i,\theta_i)\\
h_{2i}(\beta_i,\theta_i)
\end{bmatrix}^\top
\begin{bmatrix}
\x_i^\top \Delta_\beta \\
\x_i^\top \Delta_\theta
\end{bmatrix}\right)
\left(
\begin{bmatrix}
h_{1i}(\beta,\theta)\\
h_{2i}(\beta,\theta)
\end{bmatrix}^\top
\begin{bmatrix}
\x_i^\top \Delta_\beta \\
\x_i^\top \Delta_\theta
\end{bmatrix}\right)
\right],
\end{align*}
where $\x_i^\top \beta_i \in [\x_i^\top \beta, \x_i^\top \beta_0]$ and $\x_i^\top \theta_i \in [\x_i^\top \theta, \x_i^\top \theta_0]$ from MVT.
By expanding each term, we have,
\begin{align}\label{pprop1:term1}
-\textnormal{Term I} &= \E[A''(h_i(\beta_i,\theta_i) )\{ h_{1i}(\beta,\theta) h_{1i}(\beta_i,\theta_i)(\x_i^\top\Delta_\beta)^2 \})     ] \nonumber\\
&+ \E[A''(h_i(\beta_i,\theta_i) )\{ h_{2i}(\beta,\theta) h_{2i}(\beta_i,\theta_i)(\x_i^\top\Delta_\theta)^2 \})     ]  \nonumber\\
&+ \E[A''(h_i(\beta_i,\theta_i) )\{ [h_{1i}(\beta,\theta) h_{2i}(\beta_i,\theta_i) + h_{1i}(\beta_i,\theta_i) h_{2i}(\beta,\theta) ] (\x_i^\top\Delta_\beta)(\x_i^\top\Delta_\theta)\})     ].
\end{align}

For the remaining term (:= Term II) in \eqref{eq:direct_deriv},
\begin{align}\label{eq:1}
\textnormal{Term II}:=
&\E \left[(\E[ u_i z_i|\x_i] e^{-\x_i^\top \beta} - \E[u_i|\x_i] \{ 2-\sigma(\x_i^\top \beta+\log \lambda_\epsilon)  \} ) \x_i^\top \Delta_\beta\right] \nonumber \\
& = \E[\{\sigma(\x_i^\top \theta_0) \sigma(\x_i^\top \beta_0+\log \lambda_\epsilon)  e^{\x_i^\top (\beta_0-\beta)}[2- \sigma(\x_i^\top \beta_0+\log \lambda_\epsilon) ] \nonumber\\ 
& \qquad -\left. \sigma(\x_i^\top \beta_0+\log \lambda_\epsilon)\sigma(\x_i^\top \theta_0) [2- \sigma(\x_i^\top \beta+\log \lambda_\epsilon)] \} \x_i^\top \Delta_\beta\right]\nonumber\\
& = \E \left[\sigma(\x_i^\top \theta_0) \sigma(\x_i^\top \beta_0+\log \lambda_\epsilon)  e^{-\x_i^\top \beta}\{ e^{\x_i^\top \beta_0} [2- \sigma(\x_i^\top \beta_0+\log \lambda_\epsilon)]\right.\nonumber\\
&\quad \left.- e^{\x_i^\top \beta}[2- \sigma(\x_i^\top \beta+\log \lambda_\epsilon)]\}\x_i^\top\Delta_\beta   \right] 
\end{align}
where the first equality using Lemma \ref{lem:lem3}.

We define $f(t) = e^{t}[2-\sigma(t+\log\lambda_\epsilon)]$. Since the term inside the bracket in \eqref{eq:1} is
\[f(\x_i^\top\beta_0) - f(\x_i^\top\beta) = f'(\x_i^\top \beta_i)(\x_i^\top \beta_0 - \x_i^\top \beta) =- f'(\x_i^\top \beta_i') (\x_i^\top \Delta_\beta)  \]
by mean value theorem where $\x_i^\top \beta_i' \in [\x_i^\top \beta, \x_i^\top \beta_0]$,
we have,
\begin{align}\label{pprop1:term2}
\textnormal{Term II} &= -\E \left[\sigma(\x_i^\top \theta_0) \sigma(\x_i^\top \beta_0 + \log \lambda_\epsilon) e^{-\x_i^\top \beta} f'(\x_i^\top \beta_i')  (\x_i^\top \Delta_\beta)^2\right]\nonumber\\
&= -\E \left[\sigma(\x_i^\top \theta_0) \sigma(\x_i^\top \beta_0 + \log \lambda_\epsilon) e^{\x_i^\top (\beta_i'-\beta)} [1+\{1-\sigma(\x_i^\top \beta_i' + \log\lambda_\epsilon)\}^2]  (\x_i^\top \Delta_\beta)^2\right]
\end{align}
using $f'(t) = e^{t} [1+\{1-\sigma(t+\log\lambda_\epsilon)\}^2]$.

Combining Term I and II in \eqref{pprop1:term1} and \eqref{pprop1:term2},
\begin{align*}
 \langle  \triangledown \Rc(\omega) , \omega - \omega_0 \rangle
& = -(\textnormal{Term I + Term II}) \\
&= \E[A''(h_i(\beta_i,\theta_i) )\{ h_{1i}(\beta,\theta) h_{1i}(\beta_i,\theta_i)(\x_i^\top\Delta_\beta)^2 \})     ] \nonumber\\
&+ \E[A''(h_i(\beta_i,\theta_i) )\{ h_{2i}(\beta,\theta) h_{2i}(\beta_i,\theta_i)(\x_i^\top\Delta_\theta)^2 \})     ]  \nonumber\\
&+ \E[A''(h_i(\beta_i,\theta_i) )\{ [h_{1i}(\beta,\theta) h_{2i}(\beta_i,\theta_i) + h_{1i}(\beta_i,\theta_i) h_{2i}(\beta,\theta) ] (\x_i^\top\Delta_\beta)(\x_i^\top\Delta_\theta)\})     ] \nonumber\\
&+ \E \left[\sigma(\x_i^\top \theta_0) \sigma(\x_i^\top \beta_0+ \log \lambda_\epsilon)e^{\x_i^\top (\beta_i'-\beta)} [1+\{1-\sigma(\x_i^\top \beta_i' + \log\lambda_\epsilon)\}^2]   (\x_i^\top \Delta_\beta)^2\right]\nonumber.
\end{align*}
Note that other than the third terms, all other terms are positive, since $A''\geq 0$ and $0\leq h_1,h_2,\sigma,\sigma'\leq 1$. Define
\begin{align*}
g_i(\omega):= \frac{\sigma(\x_i^\top \theta_0) \sigma(\x_i^\top \beta_0+ \log \lambda_\epsilon)e^{\x_i^\top (\beta_i'-\beta)} [1+\{1-\sigma(\x_i^\top \beta_i' + \log\lambda_\epsilon)\}^2] }{A''(h_i(\beta_i,\theta_i))},
\end{align*}
and let
\begin{align}\label{eq:L}
  L:=\inf_{t,s,u,v; \max\{|t|,|s|,|u|,|v|\}\leq 2rC_X}  4\sigma(t) \sigma(s+ \log \lambda_\epsilon) e^{u} [1+(1-\sigma(v+ \log \lambda_\epsilon))^2].
\end{align}
In particular, we have $g_i(\omega) \geq L>0$ a.s. for all $i=1,\dots,n$, where $L>0$ is guaranteed by Assumption {\bfseries A3}.
We have,
\begin{align}\label{pprop1:term1_2}
  &-(\textnormal{Term I + Term II}) \nonumber \\
  &= \E[A''(h_i(\beta_i,\theta_i) )\{ h_{1i}(\beta,\theta) h_{1i}(\beta_i,\theta_i) +g_i(\omega) \}  (\x_i^\top\Delta_\beta)^2   ] + \E[A''(h_i(\beta_i,\theta_i) ) h_{2i}(\beta,\theta) h_{2i}(\beta_i,\theta_i)(\x_i^\top\Delta_\theta)^2      ]  \nonumber\\
  &\quad + \E[A''(h_i(\beta_i,\theta_i) )\{ [h_{1i}(\beta,\theta) h_{2i}(\beta_i,\theta_i) + h_{1i}(\beta_i,\theta_i) h_{2i}(\beta,\theta) ] (\x_i^\top\Delta_\beta)(\x_i^\top\Delta_\theta)\})     ] \nonumber\\
  &\geq \E[(0.5)A''(h_i(\beta_i,\theta_i) )\{ h_{1i}(\beta,\theta) h_{1i}(\beta_i,\theta_i) +g_i(\omega) \}  (\x_i^\top\Delta_\beta)^2   ]  \nonumber\\
  &\quad + \E[A''(h_i(\beta_i,\theta_i) ) \left\lbrace h_{2i}(\beta,\theta) h_{2i}(\beta_i,\theta_i)    - \frac{[h_{1i}(\beta,\theta) h_{2i}(\beta_i,\theta_i) + h_{1i}(\beta_i,\theta_i) h_{2i}(\beta,\theta) ]^2}{2( h_{1i}(\beta,\theta) h_{1i}(\beta_i,\theta_i) +g_i(\omega))}
  \right\rbrace(\x_i^\top\Delta_\theta)^2      ]
\end{align}
where we use the arithmetic inequality $ab \leq (a^2+b^2)/2$ with
\begin{align*}
  a &:= \{h_{1i}(\beta,\theta) h_{1i}(\beta_i,\theta_i) + g_i(\omega) \}^{1/2} \x_i^\top \Delta_\beta \\
  b &:= \frac{h_{1i}(\beta,\theta) h_{2i}(\beta_i,\theta_i) + h_{1i}(\beta_i,\theta_i) h_{2i}(\beta,\theta) }{ \{h_{1i}(\beta,\theta) h_{1i}(\beta_i,\theta_i) + g_i(\omega)\}^{1/2} } \x_i^\top \Delta_\theta.
\end{align*}
From the Assumption {\bfseries A4},
\begin{align*}
  &\max_{1\leq i \leq n}  \frac{1 - \sigma( \x_i^\top \beta + \log \lambda_\epsilon)}{1-\sigma (\x_i^\top \theta)}\\
  &= \max_{1\leq i \leq n}
  \frac{(1 - \sigma( \x_i^\top \beta + \log \lambda_\epsilon))/(\sigma( \x_i^\top \beta + \log \lambda_\epsilon)\sigma(\x_i^\top \theta))}{(1-\sigma (\x_i^\top \theta))/(\sigma( \x_i^\top \beta + \log \lambda_\epsilon)\sigma(\x_i^\top\theta))}\\
  &=\max_{1\leq i \leq n} \frac{h_{1i}(\beta,\theta)}{h_{2i}(\beta,\theta)}\\
  &\leq r_0(\omega_0, C_X, r)
\end{align*}
where we use the definition of $h_1$ and $h_2$ in \eqref{eq:h1_h2} and for $r_0(\omega_0, C_X, r) := (0.5)L^{1/2}$ for $L$ defined in \eqref{eq:L}.
For the ease of notation, we let $r_0 :=  r_0(\omega_0, C_X, r)$. For the part in bracket in \eqref{pprop1:term1_2}, we have,
\begin{align*}
  & h_{2i}(\beta,\theta) h_{2i}(\beta_i,\theta_i)    - \frac{[h_{1i}(\beta,\theta) h_{2i}(\beta_i,\theta_i) + h_{1i}(\beta_i,\theta_i) h_{2i}(\beta,\theta) ]^2}{2( h_{1i}(\beta,\theta) h_{1i}(\beta_i,\theta_i) +g_i(\omega))}\\
  & \leq h_{2i}(\beta,\theta) h_{2i}(\beta_i,\theta_i)    - \frac{ 2r_0^2 h_{2i}(\beta_i,\theta_i)^2  h_{2i}(\beta,\theta)^2}{g_i(\omega)}\\
   & = h_{2i}(\beta,\theta) h_{2i}(\beta_i,\theta_i)   \left\lbrace 1 - \frac{ 2r_0^2 h_{2i}(\beta_i,\theta_i)  h_{2i}(\beta,\theta)}{g_i(\omega)}\right\rbrace \\
  & \leq \frac{1}{2}h_{2i}(\beta,\theta) h_{2i}(\beta_i,\theta_i)
\end{align*}
where we use $0\leq h_1, h_2 \leq 1$, and the last inequality is from the condition $r_0^2\leq (0.25)L$.
Therefore,
\begin{align*}
  &-(\textnormal{Term I + Term II}) \\
  &\geq \E[(0.5)A''(h_i(\beta_i,\theta_i) )\{ h_{1i}(\beta,\theta) h_{1i}(\beta_i,\theta_i) +g_i(\omega) \}  (\x_i^\top\Delta_\beta)^2   ]  \nonumber\\
  &\quad + \E[(0.5)A''(h_i(\beta_i,\theta_i) )  h_{2i}(\beta,\theta) h_{2i}(\beta_i,\theta_i)    (\x_i^\top\Delta_\theta)^2      ] \\
  &\geq C_\lambda\{(C_0+L) \|\Delta_\beta\|_2^2 + C_0 \|\Delta_\theta\|_2^2\},
\end{align*}
where $\displaystyle C_0:= \inf_{t,s,u,v; \max\{|t|,|s|,|u|,|v|\}\leq 2rC_X} (0.5) A''(h(u,v) )\{ h_{1}(t,s) h_{1}(u,v) \} $, since $h_1(t,s) = h_2(s,t)$ and $g_i(\omega) \geq L$, and we use Assumption {\bfseries A2}.

We conclude,
\begin{align*}
\triangledown \Rc(\omega)^\top  (\omega-\omega_0)
\geq  C_\lambda C_0(\|\Delta_\beta\|_2^2 +\|\Delta_\theta\|_2^2 ) = C_\lambda C_0 \|\omega-\omega_0\|_2^2 ,
\end{align*}
as desired. 

\begin{lemma}\label{lem:lem3}
  \begin{align*}
  &\E[ u_i z_i|\x_i] =\sigma(\x_i^\top \theta_0) \sigma(\x_i^\top \beta_0+\log \lambda_\epsilon)  e^{\x_i^\top \beta_0}[2- \sigma(\x_i^\top \beta_0+\log \lambda_\epsilon) ]
  \end{align*}
\end{lemma}
\begin{proof}
  We have $\E[ u_i z_i|\x_i] = \E[ \1{z_i>0} z_i|\x_i] = \sigma(\x_i^\top \theta_0) \int_0^\infty y(1-e^{-\lambda_\epsilon t})\lambda_X e^{-\lambda_X y}dy $ since $y_i=z_i$ on $z_i>0$. Then,
  \begin{align*}
  \E[ u_i z_i|\x_i] &= \sigma(\x_i^\top \theta_0) \left \lbrace \int_0^\infty y\lambda_X e^{-\lambda_X y}-\int_0^\infty y\lambda_X e^{-(\lambda_X+\lambda_\epsilon) y}dy \right\rbrace \\
  & = \sigma(\x_i^\top \theta_0)\{\frac{1}{\lambda_X} - \frac{\lambda_X}{(\lambda_\epsilon+\lambda_X)^2}\}\\
  & = \sigma(\x_i^\top \theta_0) \{e^{\x_i^\top \beta_0} - \frac{1}{\lambda_\epsilon}  \sigma(\x_i^\top \beta_0 +\log \lambda_\epsilon)(1-\sigma(\x_i^\top \beta_0+\log \lambda_\epsilon))\}
  \end{align*}
  Noting that
  \begin{align*}
  e^{\x_i^\top \beta_0} &= \frac{1}{\lambda_\epsilon}
  \frac{e^{\x_i^\top \beta_0+\log \lambda_\epsilon}}{1+e^{\x_i^\top \beta_0+\log \lambda_\epsilon}}(1+e^{\x_i^\top \beta_0+\log \lambda_\epsilon})\\
  & =  \frac{1}{\lambda_\epsilon} \sigma(\x_i^\top \beta_0+\log \lambda_\epsilon)(1+e^{\x_i^\top \beta_0+\log \lambda_\epsilon}),
  \end{align*}
  we have,
  \begin{align*}
  \E[ u_i z_i|\x_i] &=  \frac{1}{\lambda_\epsilon}\sigma(\x_i^\top \theta_0) \sigma(\x_i^\top \beta_0+\log \lambda_\epsilon) \{1+e^{\x_i^\top \beta_0+\log \lambda_\epsilon} - 1+\sigma(\x_i^\top \beta_0+\log \lambda_\epsilon)\}\\
  & =  \frac{1}{\lambda_\epsilon}\sigma(\x_i^\top \theta_0) \sigma(\x_i^\top \beta_0+\log \lambda_\epsilon) e^{\x_i^\top \beta_0+\log \lambda_\epsilon} \{1+\frac{1}{1+e^{\x_i^\top \beta_0+\log \lambda_\epsilon} }\}\\
  & = \sigma(\x_i^\top \theta_0) \sigma(\x_i^\top \beta_0+\log \lambda_\epsilon)  e^{\x_i^\top \beta_0}[2- \sigma(\x_i^\top \beta_0+\log \lambda_\epsilon) ]
  \end{align*}

\end{proof}

\subsection{Proof for Theorem 2}
For the first part of Theorem 2, we perform a landscape analysis of $\L_n(\omega)$ similarly as in \citep{Mei2018-ec}. In particular, we first show that there exists an $\epsilon_0$ neigherborhood of $\omega_0$ where the population risk function $\Rc(\omega)$ is strongly convex and the gradient of $\Rc(\omega)$ does not vanish outside of the $\epsilon_0$ neighborhood. Then, using a uniform convergence result, we show that the empirical risk function $\L_n(\omega)$ has the same landscape as $\Rc(\omega)$ with high probability for a sufficiently large $n$ and therefore $\L_n(\omega)$ admits a unique stationary point inside $\epsilon_0$, which is a global minimizer of $\L_n(\omega)$ with the same high probability for a sufficiently large $n$. To establish this, we first state the following two Lemmas, whose proofs are presented at the end of section.

\begin{lemma}\label{lem:2}
\begin{enumerate}
  \item Bounds on the Hessian of $\Rc(\omega)$. There exist an $\epsilon_0>0$ and constants $0<\underline{h}<\overline{h}<\infty$ such that
  \begin{align*}
  \underline{h} \leq \inf_{\omega \in \B_2(\epsilon_0; \omega_0)} \lambda_{\rm min} (\triangledown^2 \Rc(\omega)) \leq \sup_{\omega \in \B_2(r)} \lambda_{\rm max} (\triangledown^2 \Rc(\omega))  \leq \overline{h}
  \end{align*}
  \item Bounds on the gradient of $\Rc(\omega)$. There exist constants $0<\underline{g}<\overline{g}<\infty$ such that
    \begin{align*}
  \underline{g} \leq \inf_{\omega \in \B_2(r)\setminus \B_2(\epsilon_0; \omega_0)} \|\triangledown \Rc(\omega)\|_2 \leq \sup_{\omega \in \B_2(r)} \|\triangledown \Rc(\omega)\|_2  \leq \overline{g}
  \end{align*}
\end{enumerate}
\end{lemma}

\begin{lemma}[Theorem 1 in \citealp{Mei2018-ec}]\label{lem:3} For $C = C_0 \left\lbrace\log\left(\frac{r C_Y K_X}{\delta} \right) \vee 1\right\rbrace$ for a constant $C_0$ depending on the model parameters $(C_X,r)$,
\begin{align*}
&\P\left(\sup_{\omega \in \B_2(r)} \| \triangledown \L_n(\omega) - \triangledown \Rc(\omega) \|_2 \leq C_YK_X \sqrt{\frac{C p \log n}{n}}   \right)\geq 1-\delta \\
& \P\left(\sup_{\omega \in \B_2(r)} \| \triangledown^2 \L_n(\omega) - \triangledown^2 \Rc(\omega) \|_2 \leq C_Y^2K_X^2 \sqrt{\frac{C p \log n}{n}}   \right)\geq 1-\delta
\end{align*}
given $n \geq C p\log (p)$.
\end{lemma}

Provided  Lemma \ref{lem:2} and \ref{lem:3} are true, we can choose $n \geq Cp\log n$ to be sufficiently large so that the following inequalities hold with probabilitiy $1-\delta$.
\begin{align}
&\underline{h}/2 \leq \inf_{\omega \in \B_2(\epsilon_0; \omega_0)} \lambda_{\rm min} (\triangledown^2 \L_n(\omega)) \leq \sup_{\omega \in \B_2(r)} \lambda_{\rm max} (\triangledown^2 \L_n(\omega))  \leq 2\overline{h}\label{eq:emp_ineq1}\\
&\underline{g}/2 \leq \inf_{\omega \in \B_2(r)\setminus \B_2(\epsilon_0; \omega_0)} \|\triangledown \L_n(\omega)\|_2 \leq \sup_{\omega \in \B_2(r)} \|\triangledown \L_n(\omega)\|_2  \leq 2\overline{g}\label{eq:emp_ineq2}\\
& \sup_{\omega \in \B_2(r)} \| \triangledown \L_n(\omega) - \triangledown \Rc(\omega) \|_2 \leq \frac{\alpha\epsilon_0}{4}  \label{eq:emp_ineq3}
\end{align}
We let $\mathcal{E}_n$ be the event where inequalities  \eqref{eq:emp_ineq1} - \eqref{eq:emp_ineq3} hold.

First, we argue that $\widehat{\omega}$ is an inner point of $\B_2(r)$, i.e. $\widehat{\omega} \in \B_2(r) \setminus \partial \B_2(r)$  on $\mathcal{E}_n$. To see this, suppose $\widehat{\omega} \in \partial \B_2(r)$. By the first-order optimality condition, we have,
\begin{align*}
\langle \triangledown \L_n(\widehat{\omega}) , \widehat{\omega} - \omega_0 \rangle \leq 0
\end{align*}
On the other hand,
\begin{align}\label{eq:eq14}
\langle \triangledown \L_n(\widehat{\omega}) , \widehat{\omega} - \omega_0 \rangle
&\geq \langle \triangledown \Rc(\widehat{\omega}) , \widehat{\omega} - \omega_0 \rangle - \sup_{\omega \in \B_2(r)} \| \triangledown \L_n(\omega) - \triangledown \Rc(\omega) \|_2 \|\widehat{\omega} - \omega_0\|_2 \nonumber\\
&\geq \alpha\| \widehat{\omega} - \omega_0\|_2^2 - \frac{r\alpha}{4}\|\widehat{\omega} - \omega_0\|_2 \\
&\geq  \frac{\alpha r^2}{8} \nonumber,
\end{align}
since $\| \omega_0\|_2\leq r/2$ and $\|\widehat{\omega}\|_2=r$, we have $\|\widehat{\omega} - \omega_0\|_2 \geq r/2$. We also use Proposition 1 and \eqref{eq:emp_ineq3} to obtain: $ \langle \triangledown \Rc(\widehat{\omega}) , \widehat{\omega} -\omega_0 \rangle \geq \alpha\| \widehat{\omega} - \omega_0\|_2^2 $ and  $\sup_{\omega \in \B_2(r)}\| \triangledown \L_n(\omega) - \triangledown \Rc(\omega) \|_2 \leq \alpha \epsilon_0/4\leq  \alpha r/4$.
Therefore, we have a contradiction, and conclude that $\widehat{\omega}$ is an inner point of $\B_2(r)$. Then from \eqref{eq:emp_ineq1} and \eqref{eq:emp_ineq2} we can conclude that $\L_n(\omega)$ has the unique stationary point in $\B_2(\epsilon_0; \omega_0)$ on $\mathcal{E}_n$.

Now we address the second part of the Theorem 2. On $\mathcal{E}_n$, since $\widehat{\omega}$ is an inner point of $\B_2(r)$, we have $\triangledown\L_n(\widehat{\omega})=0$.  First, we note that
\begin{align*}
& \langle \triangledown \Rc(\widehat{\omega}) , \widehat{\omega} - \omega_0 \rangle\\
&= \langle \triangledown \Rc(\widehat{\omega}) - \triangledown \L_n(\widehat{\omega} ), \widehat{\omega} - \omega_0 \rangle
+ \langle \triangledown\L_n(\widehat{\omega} ) - \triangledown \L_n( \omega_0), \widehat{\omega}-\omega_0\rangle
+ \langle \triangledown \L_n(\omega_0) - \triangledown\Rc(\omega_0), \widehat{\omega} - \omega_0 \rangle
  \end{align*}
where we use $\triangledown\L_n(\widehat{\omega}) , \triangledown\Rc(\omega_0) = 0$.

Using the Proposition 1,
\begin{align*}
    \alpha \|\widehat{\omega} - \omega_0\|_2^2 & \leq \|\triangledown \L_n (\omega_0)\|_2 \|\widehat{\omega} -\omega_0\|_2 + 2\sup_{\omega \in \B_2(r)} \|\triangledown \L_n(\omega) - \triangledown\Rc(\omega)\|_2\|\widehat{\omega} - \omega_0\|_2,
\end{align*}
and therefore
\begin{align*}
\|\widehat{\omega} - \omega_0\|_2 \leq \frac{1}{\alpha }\{ \|\triangledown \L_n (\omega_0)\|_2 +2\sup_{\omega \in \B_2(r)} \|\triangledown \L_n(\omega) - \triangledown\Rc(\omega)\|_2 \}.
\end{align*}

First, we obtain a bound on $\|\triangledown \L_n (\omega_0)\|_2 $. Using $\ell_2$-$\ell_\infty$ inequality, we have,
\begin{align*}
\|\triangledown \L_n (\omega_0)\|_2 \leq \sqrt{2p} \|\triangledown \L_n (\omega_0)\|_\infty.
\end{align*}
Now we show that $\triangledown \L_n (\omega_0)$ is sub-Gaussian with a parameter scaling with $1/n$.
We recall, for $\omega = (\beta,\theta)$,
\begin{align*}
\triangledown \L_n (\beta, \theta ) = \frac{1}{n}\sum_{i=1}^n\begin{bmatrix}
\{(A'(h_i(\beta,\theta))-u_i ) h_{1i}(\beta,\theta) -u_iz_i e^{-\x_i^\top \beta} + u_i (2-\sigma(\x_i^\top \beta + \log \lambda_\epsilon))\}\x_i\\
(A'(h_i(\beta,\theta))-u_i ) h_{2i}(\beta,\theta)  \x_i
\end{bmatrix}
\end{align*}

Let $c_{1i}(\beta,\theta) := (A'(h_i(\beta,\theta))-u_i ) h_{1i}(\beta,\theta) -u_iz_i e^{-\x_i^\top \beta} + u_i (2-\sigma(\x_i^\top \beta + \log \lambda_\epsilon))\}$ and $c_{2i} (\beta,\theta) = (A'(h_i(\beta,\theta))-u_i ) h_{2i}(\beta,\theta)$. We have $\E[c_{1i}(\beta_0,\theta_0)\x_i] ,\E[c_{2i}(\beta_0,\theta_0)|\x_i] = 0 $ by the iterative law of expectation.

Also $c_{1i}(\beta_0,\theta_0)$ and $c_{2i}(\beta_0,\theta_0)$ are bounded a.s., since $|(A'(h_i(\beta,\theta))-u_i ) h_{1i}(\beta,\theta) | , |(A'(h_i(\beta,\theta))-u_i ) h_{2i}(\beta,\theta) | \leq 1 $ almost surely, $| u_iz_i e^{-\x_i^\top \beta_0} |=| u_iy_i e^{-\x_i^\top \beta_0} | \leq  C_Y $ by Assumption {\bfseries A3}, and $|u_i (2-\sigma(\x_i^\top \beta + \log \lambda_\epsilon))| \leq 2$.

We show that $ \triangledown \L_i(\beta_0,\theta_0 )$ is a mean-zero sub-Gaussian random variable. First, we define for a random vector $\mathbf{X}$, $\|\mathbf{X}\|_{L_k(\P)} = \E[\|\mathbf{X}\|^{k}]^{1/k}$.
 For any unit vector $v \in \R^{2p}$ and $k \geq 1$
\[\|v^\top \triangledown \L_i(\beta_0,\theta_0 )\|_{L_k(\P)}  = \|c_{1i}(\beta_0,\theta_0) \x_i^\top v_\beta + c_{2i}(\beta_0,\theta_0)\x_i^\top v_\theta \|_{L_k(\P)}  \leq  C_0 C_Y K_X \sqrt{k} \]
where $C_0$ is an absolute constant and we let $v =  [v_\beta^\top,v_\theta^\top]^\top$. Therefore, $\triangledown \L_i(\beta_0,\theta_0 )$ is sub-Gaussian distribution with parameter $C_0' C_Y K_X$ for an absolute constant $C_0'$ \citep{Vershynin2018-xl}. Since $\{ \triangledown \L_i(\beta_0,\theta_0 )\}_{i=1}^n$ are independent, it follows that $\triangledown \L_n(\omega ) = n^{-1}\sum_{i=1}^n \triangledown\L_i(\omega)$ has a sub-Gaussian distribution with parameter $C_0 C_Y K_X /n $. Using a sub-Gaussian tail bound and also a union bound, for any $t>0$, we have,
\begin{align*}
  \P(\|\triangledown \L_n (\omega_0)\|_\infty \geq  t  ) \leq \exp ( -nt^2 /(2 C_0 C_Y K_X) + \log (4p))
\end{align*}
since $\omega \in \R^{2p}$.
Take $t = \sqrt{(2+(\delta\log(2p))^{-1})\log (2p) C_0C_YK_X/n}$ to get $	\P(\|\triangledown \L_n (\omega_0)\|_\infty \geq  t  )  \leq  \delta$.

The bound for the second term follows from Lemma \ref{lem:3}. Combining the two bounds, with probability $1-3\delta$ and a constant $C_1>0$,
\begin{align*}
\|\widehat{\omega} - \omega_0\|_2 &\leq \frac{1}{\alpha } \{  C_1 \sqrt{\frac{p\log p C_YK_X}{n}} +C_YK_X \sqrt{\frac{C p \log n}{n}} \}\\
&\leq \frac{C}{\alpha} \sqrt{\frac{C_Y^2 p \log (n) \log(C_Y/\delta)}{n}}.
\end{align*}

\subsection{Proof for Theorem 3}
Suppose $\mathcal{E}_n$ holds, i.e. we are on the event where inequalities \eqref{eq:emp_ineq1} - \eqref{eq:emp_ineq3} hold. On $\mathcal{E}_n$, $\L_n(\beta)$ is $2\overline{h}$-smooth and $\underline{h}/2$-strongly convex on $\B_2(\epsilon_0; \omega_0)$.

First suppose $\omega^t \in \B_2(\epsilon_0 ; \omega_0)$. We will show that iterates stay in the strongly convex region of $\B_2(\epsilon_0; \omega_0)$ and $\{\omega^k\}_{k\geq t}$ linearly converge to $\widehat{\omega}$. From the fundamental prox-grad inequality (Theorem 10.16 in \citealp{Beck2017-us}),
for any $\eta$ such that $\eta \leq 1/(2\overline{h})$,
\begin{align*}
\L_n(\widehat{\omega}) - \L_n(\omega^{t+1}) \geq \frac{1}{2\eta}\{ \|\widehat{\omega} - \omega^{t+1}\|_2^2 -\|\widehat{\omega} - \omega^{t}\|_2^2\} + \L_n(\widehat{\omega}) -\L_n(\omega^t)  -\langle \triangledown \L_n(\omega^t) , \widehat{\omega} - \omega^t\rangle
\end{align*}
Since both $\omega^t , \widehat{\omega} \in \B_2(\epsilon_0; \omega_0)$, we have,
\begin{align*}
\L_n(\widehat{\omega}) -\L_n(\omega^t)  -\langle \triangledown \L_n(\omega^t) , \widehat{\omega} - \omega^t\rangle \geq \frac{\underline{h}}{4} \|\widehat{\omega} - \omega^t\|_2^2.
\end{align*}
Therefore,
\begin{align*}
\L_n(\widehat{\omega}) - \L_n(\omega^{t+1}) \geq   \frac{1}{2\eta} \|\widehat{\omega} - \omega^{t+1}\|_2^2 - ( \frac{1}{2\eta}- \frac{\underline{h}}{4})\|\widehat{\omega} - \omega^{t}\|_2^2
\end{align*}
Since $\widehat{\omega}$ is the unique minimizer of $\L_n(\omega)$ on $\mathcal{E}_n$, we have $\L_n(\widehat{\omega}) \leq  \L_n(\omega^{t+1}) $, and thus
\begin{align}\label{pthm3:ineq1}
\|\widehat{\omega} - \omega^{t+1}\|_2^2  \leq (1- \frac{\underline{h}\eta}{2})\|\widehat{\omega} - \omega^{t}\|_2^2
\end{align}
Since $\eta \leq 1/(2\overline{h})$, $\frac{\underline{h}\eta}{2} \leq \frac{\underline{h}}{4\overline{h}} <1$. Therefore, if $\omega^t \in \B_2(\epsilon_0 ; \omega_0)$, then $\omega^{t+1} \in \B_2(\epsilon_0 ; \omega_0)$ and $\{\omega^k\}_{k\geq t}$ linearly converge to $\widehat{\omega}$.

Now suppose $\omega^t \notin \B_2(\epsilon_0 ; \omega_0)$. Similar arguments as in \cite{Mei2018-ec} for the analysis of the gradient descent algorithm can be used to show that iterates exponentially converge to the strongly convex region. Since the iterates do not leave $\B_2(\epsilon_0 ; \omega_0)$ once the iterates enter this region, we can assume that $\omega^k \notin \B_2(\epsilon_0 ; \omega_0)$ for $k =0,\dots,t$. We have,
\begin{align*}
\|\omega^{t+1} - \omega_0\|_2^2
&= \|\mathcal{P}_{\B_2(r)}(\omega^{t}-\eta\triangledown\L_n(\omega^t) )-\mathcal{P}_{\B_2(r)} (\omega_0)\|_2^2\\
&\leq \| \omega^{t}-\eta\triangledown\L_n(\omega^t) - \omega_0\|_2^2\\
&\leq \| \omega^{t}- \omega_0\|_2^2 - 2\eta \langle \omega^{t}- \omega_0, \triangledown \L_n(\omega^t) \rangle +\eta^2 \|\triangledown\L_n(\omega^t)\|_2^2
\end{align*}
where the first inequality uses the contraction property of a projection operator.
We have
\begin{align*}
\langle  \triangledown \L_n(\omega^t) , \omega^{t}- \omega_0\rangle
&\geq \alpha \|\omega^t - \omega_0\|_2^2 - (\alpha\epsilon_0/4)  \|\omega^t - \omega_0\|_2\\
&\geq (3\alpha/4) \|\omega^t - \omega_0\|_2^2
\end{align*}
where the first inequality can be derived similarly as in \eqref{eq:eq14} and the second inequality is due to $ \|\omega^t - \omega_0\|_2 \geq \epsilon_0$.

Let $\eta \leq \frac{3}{16\overline{g}^2}\alpha\epsilon_0^2$. Then,
\begin{align*}
\|\omega^{t+1} - \omega_0\|_2^2
&= \|\mathcal{P}_{\B_2(r)}(\omega^{t}-\eta\triangledown\L_n(\omega^t) )-\mathcal{P}_{\B_2(r)} (\omega_0)\|_2^2\\
&\leq (1-3\eta\alpha/2)\| \omega^{t}- \omega_0\|_2^2  +4\eta^2 \overline{g}^2\\
&\leq (1-3\eta\alpha/4)\| \omega^{t}- \omega_0\|_2^2
\end{align*}
where we use \eqref{eq:emp_ineq2} for the second inequality and $4\eta^2 \overline{g}^2 \leq 3\eta\alpha\epsilon_0^2/4$ by the choice of $\eta.$ Then, since
$\| \widehat{\omega} - \omega_0\|_2 \leq \epsilon_0 \leq \| \omega^{t}- \omega_0\|_2 $,
\begin{align*}
\| \omega^{t}- \widehat{\omega}\|_2
&\leq \| \omega^{t}- \omega_0\|_2 +\| \widehat{\omega} - \omega_0\|_2 \\
&\leq 2\| \omega^{t}- \omega_0\|_2 \\
&\leq 2(1-3\eta\alpha/4)^{t/2}\| \omega^{0}- \omega_0\|_2 \\
&\leq 2(1-3\eta\alpha/4)^{t/2}\{\| \omega^{0}- \widehat{\omega} \|_2+\| \widehat{\omega} - \omega_0\|_2 \}\\
&\leq 4(1-3\eta\alpha/4)^{t/2}\| \omega^{0}- \widehat{\omega} \|_2
\end{align*}
where the last inequality is due to $\| \widehat{\omega} - \omega_0\|_2 \leq \epsilon_0 \leq \| \omega^{0}- \widehat{\omega} \|_2$. Therefore,
\begin{align}\label{pthm3:ineq2}
\| \omega^{t}- \widehat{\omega}\|_2^2 \leq 16(1-3\eta\alpha/4)^t \| \omega^{0}- \widehat{\omega} \|_2^2.
\end{align}

Combining \eqref{pthm3:ineq1} and \eqref{pthm3:ineq2},
\begin{align*}
\| \omega^{t}- \widehat{\omega}\|_2^2 \leq 16(1-\kappa)^t \| \omega^{0}- \widehat{\omega} \|_2^2.
\end{align*}
for $\kappa:=\min \{\underline{h},  \frac{3\alpha}{4}\}\eta$ where $\eta$ is chosen so that $\displaystyle \eta \leq  \min\{ \frac{1}{2\overline{h}}, \frac{3\alpha\epsilon_0^2}{16\overline{g}^2}, \frac{2}{3\alpha} \}$.

\begin{proof}[Proof of Lemma 2]
First, we compute Hessian of $\Rc(\omega)$:
\begin{align*}
\triangledown^2 \Rc(\omega ) = \begin{bmatrix}
\E[a_i (\omega)|\x_i] \x_i\x_i^\top & \E[b_i(\omega) |\x_i]  \x_i\x_i^\top\\
\E[b_i (\omega)|\x_i]   \x_i\x_i^\top & \E[c_i (\omega)|\x_i] \x_i\x_i^\top\\
\end{bmatrix}
\end{align*}
where
\begin{equation}\label{def:abc}
  \begin{aligned}
a_i(\omega)&:=a(\x_i^\top \beta, \x_i^\top \theta):= A''(h_i(\omega )) h_{1i} (\omega)^2 +(A'(h_i(\omega)) -u_i )h_{11,i}(\omega) \\
&\qquad\qquad\qquad\qquad+u_iz_i e^{-\x_i^\top \beta} -u_i \sigma'(\x_i^\top \beta + \log \lambda_\epsilon)\\
b_i(\omega)&:=b(\x_i^\top \beta, \x_i^\top \theta):=A''(h_i(\omega )) h_{1i} (\omega) h_{2i} (\omega) +(A'(h_i(\omega)) -u_i )h_{12,i}(\omega)\\
c_i(\omega)&:=c(\x_i^\top \beta, \x_i^\top \theta):=A''(h_i(\omega )) h_{2i} (\omega)^2 + (A'(h_i(\omega)) -u_i )h_{22,i}(\omega).
\end{aligned}
\end{equation}

First we show that $ \inf_{\omega \in \B_2(\epsilon_0; \omega_0)} \lambda_{\rm min} (\triangledown^2 \Rc(\omega)) > C $ for $C>0$.
For any $v \in \R^{2p}$ such that $\|v\|_2=1$ and $v = [v_\beta^\top,v_\theta^\top]^\top$,
\begin{align*}
v^\top \triangledown^2 \Rc(\omega_0)v
&= \E[\E[a_i(\omega_0)|\x_i](\x_i^\top v_\beta)^2+\E[c_i(\omega_0)|\x_i](\x_i^\top v_\theta)^2+2\E[b_i(\omega_0)|\x_i](\x_i^\top v_\beta)(\x_i^\top v_\theta)].
\end{align*}
We note
\begin{align*}
\E[a_i(\omega_0)|\x_i]&:= A''(h_i(\omega_0 )) h_{1i} (\omega_0)^2 +\E[u_iz_i|\x_i] e^{-\x_i^\top \beta_0} - \E[u_i|\x_i] \sigma'(\x_i^\top \beta_0 + \log \lambda_\epsilon)\\
& = A''(h_i(\omega_0 )) h_{1i} (\omega_0)^2 +\sigma(\x_i^\top \theta_0)\sigma(\x_i^\top \beta_0 + \log \lambda_\epsilon) [1+\{1-\sigma(\x_i^\top \beta_0 +\log\lambda_\epsilon)\}^2]\\
\E[b_i(\omega_0)|\x_i]&:=A''(h_i(\omega_0 )) h_{1i} (\omega_0) h_{2i} (\omega_0) \\
\E[c_i(\omega_0)|\x_i]&:=A''(h_i(\omega_0 )) h_{2i} (\omega_0)^2.
\end{align*}
where the first equality uses Lemma \ref{lem:lem3}.
Letting
\[\tilde{g}_i(\omega_0) := \frac{1}{A''(h_i(\omega_0))}\sigma(\x_i^\top \theta_0)\sigma(\x_i^\top \beta_0 + \log \lambda_\epsilon) [1+\{1-\sigma(\x_i^\top \beta_0 +\log\lambda_\epsilon)\}^2],\]
similarly as in the proof of Proposition 1,
\begin{align*}
v^\top \triangledown^2 \Rc(\omega_0)v
&\geq \E[A''(h_i(\omega_0 ))  \left[  (0.5) \{h_{1i} (\omega_0)^2+\tilde{g}_i(\omega_0)\}(\x_i^\top v_\beta)^2 +\right.\\
&\qquad\qquad\qquad\qquad\qquad\qquad\left\lbrace h_{2i} (\omega_0)^2-\frac{ 4h_{1i}(\omega_0)^2 h_{2i}(\omega_0)^2 }{2(h_{1i}(\omega_0)+\tilde{g}_i(\omega_0))} \right\rbrace(\x_i^\top v_\theta)^2 ]]\\
&\geq  \E[A''(h_i(\omega_0 ))  \left[  (0.5) \{h_{1i} (\omega_0)^2+\tilde{g}_i(\omega_0)\}(\x_i^\top v_\beta)^2 +
h_{2i} (\omega_0)^2\left\lbrace 1-\frac{ 2r_0^2  }{\tilde{g}_i(\omega_0)} \right\rbrace(\x_i^\top v_\theta)^2 \right]]\\
&\geq\E[A''(h_i(\omega_0 ))  \left[  (0.5) \{h_{1i} (\omega_0)^2+\tilde{g}_i(\omega_0)\}(\x_i^\top v_\beta)^2 +
(0.5) h_{2i} (\omega_0)^2 (\x_i^\top v_\theta)^2 \right]]
\end{align*}
where we use $L\geq \tilde{g}_i(\omega_0)$ for all $i$ a.s., and Assumption {\bfseries A4} for $r_0 := (0.5)L^{1/2}$. Therefore,
\begin{align*}
v^\top \triangledown^2 \Rc(\omega_0)v  \geq C_0  \E[ (\x_i^\top v_\beta)^2+(\x_i^\top v_\theta)^2] \geq 2C_0C_\lambda
\end{align*}
for $\displaystyle C_0:= \inf_{t,s ; \max\{|t|,|s|\}\leq 2rC_X} (0.5) A''(h(s,t) )h_{1}(s,t)^2   $, since $h_1(t,s) = h_2(s,t)$, where we also use Assumption {\bfseries A2}.
Thus we obtain $\lambda_{\rm min}(\triangledown^2 \Rc(\omega_0)) \geq 2C_0C_\lambda$. To bound $\lambda_{\rm min}(\triangledown^2 \Rc(\omega))$, we use
\begin{align*}
\lambda_{\rm min}(\triangledown^2 \Rc(\omega)) \geq \lambda_{\rm min}(\triangledown^2 \Rc(\omega_0)) - \|\triangledown^2 \Rc(\omega ) - \triangledown^2 \Rc(\omega_0) \|_{op}
\end{align*}

For any $\omega_1$ and $\omega_2$ we have,
\begin{align*}
\|\triangledown^2 \Rc(\omega_1 ) - \triangledown^2 \Rc(\omega_2) \|_{op}
& =
\| \begin{bmatrix}
\E[\E[a_i (\omega_1)-a_i (\omega_2)|\x_i] \x_i\x_i^\top] &\E[\E[b_i (\omega_1)-b_i (\omega_2)|\x_i]\x_i\x_i^\top]\\
\E[\E[b_i (\omega_1)-b_i (\omega_2)|\x_i]\x_i\x_i^\top] & \E[\E[c_i (\omega_1)-c_i (\omega_2)|\x_i]\x_i\x_i^\top]\\
\end{bmatrix}\|_{op}\\
& =
\sup_{v; \|v\|_2=1}  \E\left[\E[a_i (\omega_1)-a_i (\omega_2)|\x_i] (\x_i^\top v_\beta)^2
+ 2\E[b_i (\omega_1)-b_i (\omega_2)|\x_i](\x_i^\top v_\beta) (\x_i^\top v_\theta )\right.\\
& \left.\qquad + \E[c_i (\omega_1)-c_i (\omega_2)|\x_i] (\x_i^\top v_\theta )^2\right]\\
&\leq \E\{L_a\|\x_i\|_2 (\x_i^\top v_\beta)^2  + L_c \|\x_i\|_2 (\x_i^\top v_\theta)^2 + 2L_b \|\x_i\|_2|\x_i^\top \beta \x_i^\top \theta|\}\|\omega_1-\omega_2\|_2 \\
&\leq C_X^3(L_a+2L_b+L_c)\|\omega_1-\omega_2\|_2
\end{align*}
where $L_a$, $L_b$, and $L_c$ are Lipschitz constants from Lemma \ref{lem:bound_abc}, and for the first inequality we use Lipschitz and H\"older inequalities in a way that $|a_i(\omega_1)  - a_i(\omega_2) | =|a(\x_i^\top \beta_1,\x_i^\top \theta_1)  - a(\x_i^\top \beta_2,\x_i^\top \theta_2) | \leq L_a\|\x_i^\top [\omega_1- \omega_2]\|_2 \leq L_a \|\x_i\|_2 \|\omega_1-\omega_2\|_2$. Therefor for any $\omega$ such that $\|\omega-\omega_0\|_2\leq \epsilon_0$, for
\begin{align*}
\epsilon_0 := \frac{C_0C_\lambda}{2C_X^3(L_a+2L_b+L_c)},
\end{align*}
we have,
\begin{align*}
\lambda_{\rm min}(\triangledown^2 \Rc(\omega)) \geq C_0C_\lambda
\end{align*}
and we let $  \underline{h} :=C_0C_\lambda$.
For the upper bound of $\triangledown^2 \Rc(\omega)$,
\begin{align*}
\| \triangledown^2 \Rc(\omega )\|_2 &= \sup_{v; \|v\|_2 =1} v^\top \triangledown^2 \Rc(\omega )  v \\
&=\E[a_i(\omega)(v_\beta^\top \x_i)^2 + c_i (\omega) (v_\theta^\top \x_i)^2 + 2 b_i(\omega) (v_\beta ^\top \x_i )(v_\theta^\top \x_i)]\\
& \leq \E[a_i(\omega)(v_\beta^\top \x_i)^2 + c_i (\omega) (v_\theta^\top \x_i)^2 + 2 |b_i(\omega) (v_\beta ^\top \x_i )(v_\theta^\top \x_i)|].
\end{align*}

Using Lemma \ref{lem:bound_abc} and H\"older's inequality,
\begin{align*}
\| \triangledown^2 \Rc(\omega )\|_2 & \leq C_a \E[(v_\beta^\top \x_i)^2] + 2E[ (v_\theta^\top \x_i)^2] + 4 \E[|v_\beta ^\top \x_i | |v_\theta^\top \x_i|]\\
&\leq (C_a+6)C_X^2,
\end{align*}
where $C_a: = 3+C_Ye^{C_Xr}$. Therefore we can take $\overline{h} =  (C_a+6)C_X^2$.

Now we address bounds on gradients. For the lower bound, we can use Proposition 1 and Cauchy-Schuwarz inequality to obtain
\begin{align*}
\|\triangledown \Rc(\omega)\|_2\|\omega-\omega_0\|_2 \geq \alpha \|\omega-\omega_0\|_2^2.
\end{align*}
Therefore for $\|\omega-\omega_0\|_2\geq \epsilon_0$, $\|\triangledown \Rc(\omega)\|_2 \geq \epsilon_0 \alpha$. We can set $\underline{g} := \epsilon_0 \alpha$.
Finally, for the upper bound of $\|\triangledown \Rc(\omega)\|_2$, we have,
\begin{align*}
\triangledown \Rc(\omega ) = \begin{bmatrix}
\E[ (\{A'(h_i(\omega))-u_i\} h_{1i}(\omega) +u_i\{- e^{-\x_i^\top \beta}z_i +2-\sigma(\x_i^\top \beta+\log \lambda_\epsilon) \} ) \x_i]\\
\E[\{A'(h_i(\omega))-u_i\} h_{2i}(\omega) \x_i]
\end{bmatrix}.
\end{align*}
Therefore,
\begin{align*}
\|\triangledown \Rc(\omega ) \|_2
&=\sup_{v; \|v\|_2 =1} \triangledown \Rc(\omega )^\top v\\
& = \E[\{A'(h_i(\omega))-u_i\} h_{1i}(\omega) \x_i^\top v_\beta + \{A'(h_i(\omega))-u_i\} h_{2i}(\omega) \x_i^\top v_\theta \\
& \qquad +\{ \E[u_i|\x_i] (2-\sigma(\x_i^\top \beta+\log \lambda_\epsilon ))   - \E[u_iz_i|\x_i] e^{-\x_i^\top \beta}\}\x_i^\top v_\beta ]
\end{align*}
Noting $|A'(h_i(\omega))-u_i|\leq 1$, for all $i$, $0 \leq h_{1}, h_2 \leq 1$, and also using Lemma \ref{lem:lem3},
\begin{align*}
\|\triangledown \Rc(\omega ) \|_2
& \leq \E[|\x_i^\top v_\beta|+ | \x_i^\top v_\theta |\\
& \quad + \sigma(\x_i^\top \beta_0+\log \lambda_\epsilon) \sigma(\x_i^\top \theta_0) \{
(2-\sigma(\x_i^\top \beta+\log \lambda_\epsilon ))  \\
&\quad - e^{-\x_i^\top (\beta_0-\beta)} (2-\sigma(\x_i^\top \beta_0+\log \lambda_\epsilon )) \}\x_i^\top v_\beta ]\\
&\leq  \E[(3+2e^{2C_Xr}) )|\x_i^\top v_\beta|+ | \x_i^\top v_\theta | ]\\
&\leq 2(3+2e^{2C_Xr}) C_X,
\end{align*}
and we can set $\overline{g}:= 2(3+2e^{2C_Xr}) C_X$.
\end{proof}

\begin{proof}[Proof for Lemma \ref{lem:3}]
    We verify Assumptions 1-3 in \citet{Mei2018-ec}. The first assumption is to verify whether the gradient of the loss has a sub-Gaussian tail. The second assumption is to show that the Hessian evaluated on a unit vector is sub-Exponential. The third assumption is about the Lipschitz continuity of the Hessian. We mainly check whether quantities in interest satisfy a sub-gaussian/exponential moment bounds.
    We recall, for $\omega = (\beta,\theta)$,
  \begin{align*}
  \triangledown \L_n (\beta, \theta ) = \frac{1}{n}\sum_{i=1}^n\begin{bmatrix}
  \{(A'(h_i(\beta,\theta))-u_i ) h_{1i}(\beta,\theta) -u_iz_i e^{-\x_i^\top \beta} + u_i (2-\sigma(\x_i^\top \beta + \log \lambda_\epsilon))\}\x_i\\
  (A'(h_i(\beta,\theta))-u_i ) h_{2i}(\beta,\theta)  \x_i
  \end{bmatrix}
  \end{align*}

  Let $c_{1i}(\beta,\theta) := (A'(h_i(\beta,\theta))-u_i ) h_{1i}(\beta,\theta) -u_iz_i e^{-\x_i^\top \beta} + u_i (2-\sigma(\x_i^\top \beta + \log \lambda_\epsilon))\}$ and $c_{2i} (\beta,\theta) = (A'(h_i(\beta,\theta))-u_i ) h_{2i}(\beta,\theta)$.
  Similarly as in the proof of Theorem 2, $|c_{1i}(\beta,\theta)|\leq 3+C_Ye^{C_Xr} , |c_{2i}(\beta,\theta)|\leq 1$ a.s. $\forall \omega$.  Therefore, each $\triangledown \L_i (\omega)$ is $C_1C_Ye^{C_Xr}K_X$ sub-Gaussian.

  We now check Hessian:
  \begin{align*}
  \triangledown^2 \L_i(\omega ) = \begin{bmatrix}
  a_i (\omega) \x_i\x_i^\top & b_i (\omega)\x_i\x_i^\top\\
  b_i (\omega) \x_i\x_i^\top & c_i (\omega)\x_i\x_i^\top\\
  \end{bmatrix}
  \end{align*}
  for $a_i(\omega), b_i(\omega)$, and $c_i(\omega)$ defined in \eqref{def:abc}. We show for any $k\geq 1$, $\|v^\top \triangledown^2 \L_i(\omega )  v \|_{L_k (\P)} \leq C k$ for some $C<\infty$.
  \begin{align*}
  v^\top \triangledown^2 \L_i(\omega )  v
  &=a_i(\omega)(v_\beta^\top \x_i)^2 + c_i (\omega) (v_\theta^\top \x_i)^2 + 2 b_i(\omega) (v_\beta ^\top \x_i )(v_\theta^\top \x_i)\\
  & \leq a_i(\omega)(v_\beta^\top \x_i)^2 + c_i (\omega) (v_\theta^\top \x_i)^2 + 2 |b_i(\omega) (v_\beta ^\top \x_i )(v_\theta^\top \x_i)|
  \end{align*}

  and
  \begin{align*}
  \|v^\top \triangledown^2 \L_i(\omega )  v \|_{L_k (\P)}
  &= \|a_i(\omega)(v_\beta^\top \x_i)^2 + c_i (\omega) (v_\theta^\top \x_i)^2 + 2 b_i(\omega) (v_\beta ^\top \x_i )(v_\theta^\top \x_i) \|_{L_k (\P)}\\
  & \leq \|a_i(\omega)(v_\beta^\top \x_i)^2\|_{L_k (\P)} +\| c_i (\omega) (v_\theta^\top \x_i)^2 \|_{L_k (\P)}+ 2\| b_i(\omega) (v_\beta ^\top \x_i )(v_\theta^\top \x_i) \|_{L_k (\P)}\\
  & \leq (3+C_Y e^{C_Xr}) \E[(v_\beta^\top \x_i)^{2k}]^{1/k} + 2 \E[(v_\theta^\top \x_i)^{2k}]^{1/k} + 4\E[(v_\beta^\top \x_i)^{2k}]^{1/2k} \E[(v_\theta^\top \x_i)^{2k}]^{1/2k}\\
  & \leq C_1 C_Y e^{C_Xr} K_X^2 k
  \end{align*}
  for an absolute constant $C_1>0$. For the ifrst inequality, we use the Minkowski's inequality, and we use Lemma \ref{lem:bound_abc} for the third inequality to bound $a_i(\omega), b_i(\omega)$, and $c_i(\omega)$.

  Lastly, we address the Lipschitz continuity of the Hessian. First we bound $\|\triangledown^2 \Rc(\omega_0) \|_{op}$:
  \begin{align*}
  \triangledown^2 \Rc(\omega_0)  = \begin{bmatrix}
  \E[a_i (\omega_0) \x_i\x_i^\top] & \E[b_i (\omega_0)\x_i\x_i^\top]\\
  \E[b_i (\omega_0) \x_i\x_i^\top] & \E[c_i (\omega_0)\x_i\x_i^\top]\\
  \end{bmatrix}
  \end{align*}
  we have
  \begin{align*}
  \|\triangledown^2 \Rc(\omega_0) \|_{op}
  &= \sup_{v; \|v\|_2 =1}\E [a_i(\omega_0)(v_\beta^\top \x_i)^2 + c_i (\omega_0) (v_\theta^\top \x_i)^2 + 2 b_i(\omega_0) (v_\beta ^\top \x_i )(v_\theta^\top \x_i)]\\
  & \leq C_2 C_Y  K_X^2,
  \end{align*}
  for some absolute constant $C_2>0$, and
  \begin{align*}
  J(\x_i,y_i):=&\frac{\|\triangledown^2 \L_i(\omega_1 ) - \triangledown^2 \L_i(\omega_2) \|_{op}}{\|\omega_1 - \omega_2\|_2} \\
  & =
  \| \begin{bmatrix}
  \{a_i (\omega_1)-a_i (\omega_2)\} \x_i\x_i^\top & \{b_i (\omega_1)-b_i (\omega_2)\}\x_i\x_i^\top\\
  \{b_i (\omega_1)-b_i (\omega_2)\}\x_i\x_i^\top & \{c_i (\omega_1)-c_i (\omega_2)\}\x_i\x_i^\top\\
  \end{bmatrix}\|_{op} \frac{1}{{\|\omega_1 - \omega_2\|_2} }\\
  & =  \frac{1}{{\|\omega_1 - \omega_2\|_2} }   \left[
  \sup_{v; \|v\|_2=1}  \{a_i (\omega_1)-a_i (\omega_2)\} (\x_i^\top v_\beta)^2
  + 2\{b_i (\omega_1)-b_i (\omega_2)\} (\x_i^\top v_\beta) (\x_i^\top v_\theta )\right.\\
  & \left.\qquad + \{c_i (\omega_1)-c_i (\omega_2)\} (\x_i^\top v_\theta )^2\right]\\
  &\leq L_a\|\x_i\|_2 (\x_i^\top v_\beta)^2  + L_c \|\x_i\|_2 (\x_i^\top v_\theta)^2 + 2L_b \|\x_i\|_2|\x_i^\top \beta \x_i^\top \theta|\\
  &\leq (L_a+2L_b+L_c)\|\x_i\|_2^3
  \end{align*}
  where we use Lemma \ref{lem:bound_abc} to bound e.g. $|a_i (\omega_1)-a_i (\omega_2)| \leq \|\triangledown a(s,t) \|_2 \|\begin{bmatrix}
  \x_i^\top (\beta_1-\beta_2)\\ \x_i^\top (\theta_1 - \theta_2))
  \end{bmatrix}\|_2 \leq L_a \|\x_i\| \|\omega_1 - \omega_2\|_2$.
  
  Therefore,
  $\E[J(\x_i,y_i)] \leq C_5 C_Ye^{C_Xr} K_X^3 p^{3/2}$ for some absolute constant $C_5>0$,
  since $\E[\|\x\|_2^3] = \E[(\sum_{i=1}^p x_i^2)^{3/2}] \leq p^{1/2} \E[(\sum_{i=1}^p |x_i|^3)] \leq 3^{3/2} K_X^3 p^{3/2}$.
\end{proof}

\begin{lemma}\label{lem:bound_abc}
For any $\omega \in B_2(r)$, and (random) functions $a,b,c:\R\times\R \rightarrow \R$ defined as
\begin{align*}
a(t, s) &:= A''(h(t, s )) h_{1} (t, s)^2 +(A'(h(t, s)) -u_i )h_{11}(t, s) +u_iz_i e^{-t} -u_i \sigma'(t + \log \lambda_\epsilon)\\
b(t, s) &:=A''(h(t, s )) h_{1} (t, s) h_{2} (t, s) +(A'(h_i(\omega)) -u_i )h_{12,i}(\omega)\\
c(t, s) &:=A''(h(t, s )) h_{2} (t, s)^2 + (A'(h(t, s)) -u_i )h_{22}(t, s)
\end{align*}
we have,
\begin{align*}
|a(t,s)| \leq 3  + C_Y e^{C_Xr} ,\quad  |b(t,s)| ,|c(t,s)| \leq 2, \quad \forall t,s \quad a.s.,
\end{align*}
and
\begin{align*}
\| \triangledown a(t,s)\|_2 \leq C_1 + C_Ye^{C_Xr}, \| \triangledown b(t,s)\|_2,   \| \triangledown c(t,s)\|_2 \leq C_2 \quad a.s,
\end{align*}
where $C_1$ and $C_2$ are some absolute constants.
\end{lemma}
\begin{proof}
First, $A''(t) \leq 0.25, \forall t$. We also have $h_2(t,s) = h_1(s,t)$. Therefore we only need to compute $h_{11}$ and $h_{12}$ for terms involving second derivatves of $h$ since $h_{22}(t,s) = h_{11}(s,t)$. From direct computation,
\begin{align*}
h_{11}(t,s) &= -\frac{e^t(1+e^s)}{(1+e^s+e^t)^2}\\
h_{12}(t,s) &= \frac{e^{t+s}}{(1+e^s+e^t)^2}
\end{align*}
In particular, $\max\{|h_{11}|, |h_{22}|, |h_{12}|\} \leq 1, \forall t,s$.
Therefore,
\begin{align*}
|a(t,s)| \leq 3  + C_Y e^{C_Xr} ,\quad  |b(t,s)| ,|c(t,s)| \leq 2, \quad \forall  t,s
\end{align*}
using Assumption {\bfseries A3}.  We can also bound $\| \triangledown a(t,s)\|_2, \| \triangledown b(t,s)\|_2$, and $\| \triangledown c(t,s)\|_2$ similarly. Since each $\| \triangledown a(t,s)\|_2, \| \triangledown b(t,s)\|_2$, and $\| \triangledown c(t,s)\|_2$ have terms involving $A'''$ and third order partial derivatives of $h$, we need to bound $A'''$ and third order partial derivatives of $h$. Since $h_1(t,s) = h_2(s,t)$, we compute bounds for $h_{111}, h_{112}$ and $h_{122}$.

From direct calculation, we can obtain,
\begin{align*}
h_{111}(t,s) &= \frac{e^{2t} (1+e^s)}{(1+e^t+e^s)^3}-\frac{e^{t} (1+e^s)^2}{(1+e^t+e^s)^3}\\
h_{112}(t,s) &= \frac{e^{t}e^s (1+e^s)}{(1+e^t+e^s)^3}-\frac{e^{2t} e^s}{(1+e^t+e^s)^3}\\
h_{122}(t,s) &= \frac{e^{t}e^s }{(1+e^t+e^s)^3}-\frac{e^{2s+t} }{(1+e^t+e^s)^3}+\frac{e^{s+2t}}{(1+e^t+e^s)^3}\\
\end{align*}
In particular, they are all bounded by $3$. We have $\| \triangledown a(t,s)\|_2 \leq C_1 + C_Ye^{C_Xr}$,  \\
$\| \triangledown b(t,s)\|_2,\| \triangledown c(t,s)\|_2 \leq C_2$ where $C_1$ and $C_2$ are some absolute constants.
\end{proof}
\newpage
\section{Supplementary Figures}
\begin{figure}[htb]
\centering
\begin{subfigure}{1\textwidth}
  \centering
  \includegraphics[width=.8\linewidth]{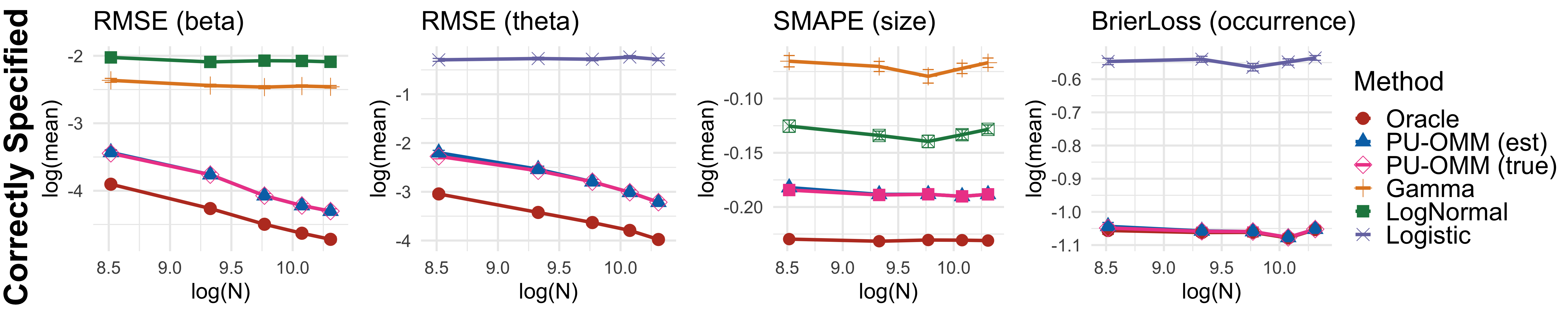}
  \label{fig:setting1-a}
\end{subfigure}
\begin{subfigure}{1\textwidth}
  \centering
  \includegraphics[width=.8\linewidth]{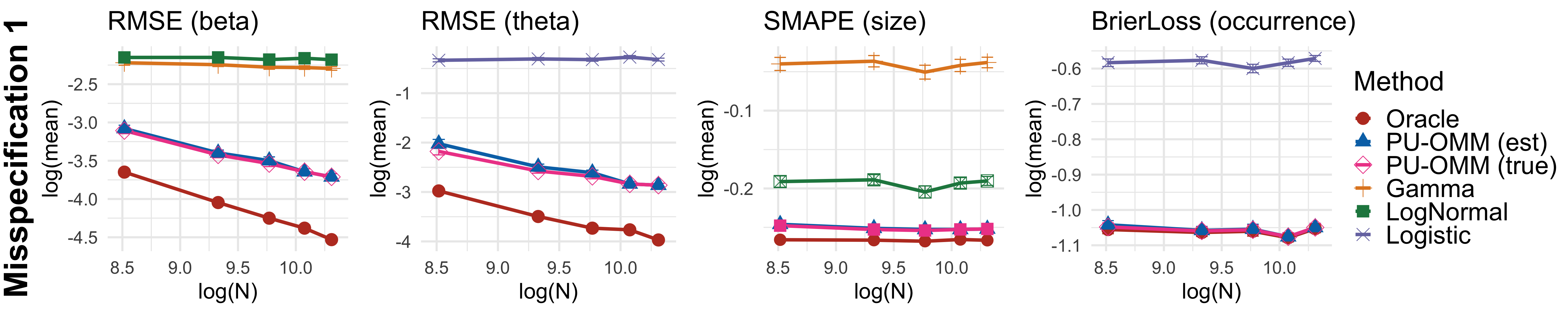}
  \label{fig:setting2-a}a
\end{subfigure}
\begin{subfigure}{1\textwidth}
\centering
\includegraphics[width=.8\linewidth]{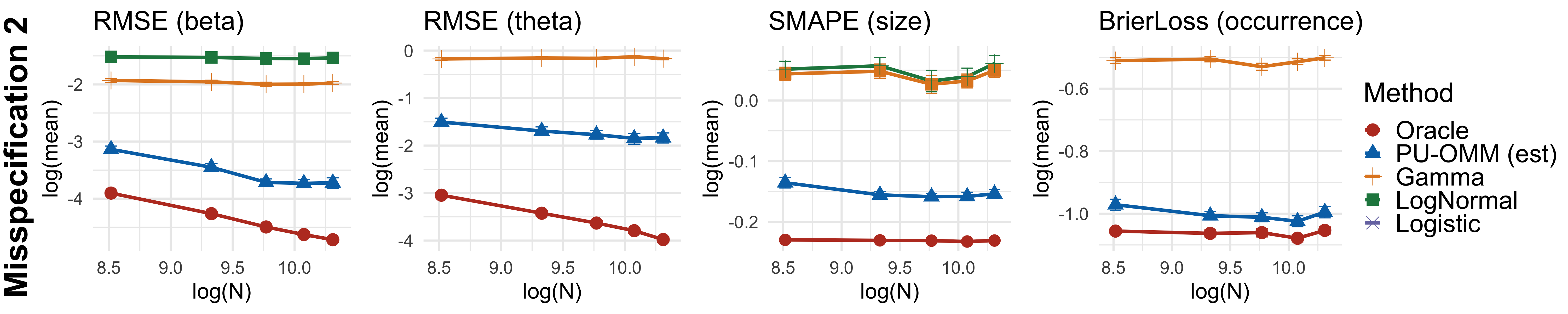}
\label{fig:setting3-a}
\end{subfigure}
\caption{Parametric estimation and prediction accuracy of Oracle, PU-OMM (est), PM-OMM (true), LogNormal-Logistic, LogNormal-Gamma under Settings 1-3. PU-OMM (true) refers to a  PU-OMM method where true $\lambda_\epsilon$ (the value that was used in data generating process) is used and PU-OMM (est) refers to a PU-OMM method where $\lambda_\epsilon$ is chosen based on goodness of fit of the observed occurrence (also see Implementation Details in the main paper). PU-OMM(true) is dropped in Setting 3 (Misspecification 2) because the true data generating process does not involve $\lambda_\epsilon$. The hyperparameter $\lambda_\epsilon$ chosen based on the goodness of fit of the observed occurrence was typically close to the true $\lambda_\epsilon$, and therefore, the performances of the two PU-OMM methods were very similar.}\label{fig:evalModels}
\end{figure}

\begin{figure*}[htb]
\centering
\begin{subfigure}{1\textwidth}
  \centering
  \includegraphics[width=.8\linewidth]{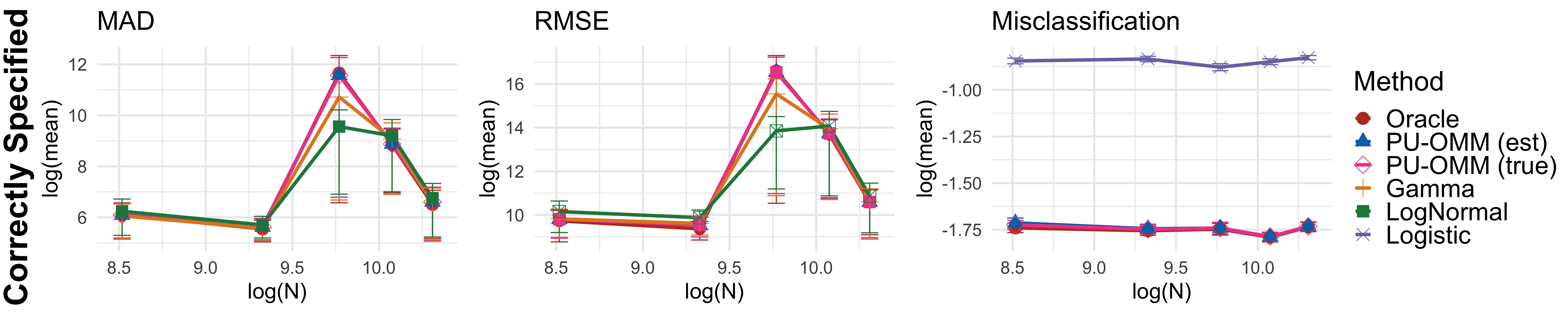}
  \label{fig:setting1-2}
\end{subfigure}
\begin{subfigure}{1\textwidth}
  \centering
  \includegraphics[width=.8\linewidth]{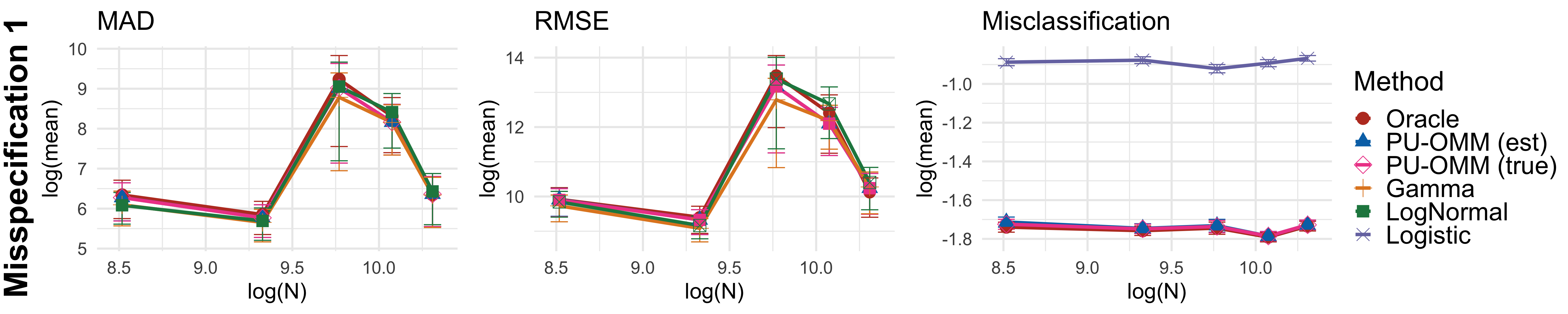}
  \label{fig:setting2-2}
\end{subfigure}
\begin{subfigure}{1\textwidth}
  \centering
  \includegraphics[width=.8\linewidth]{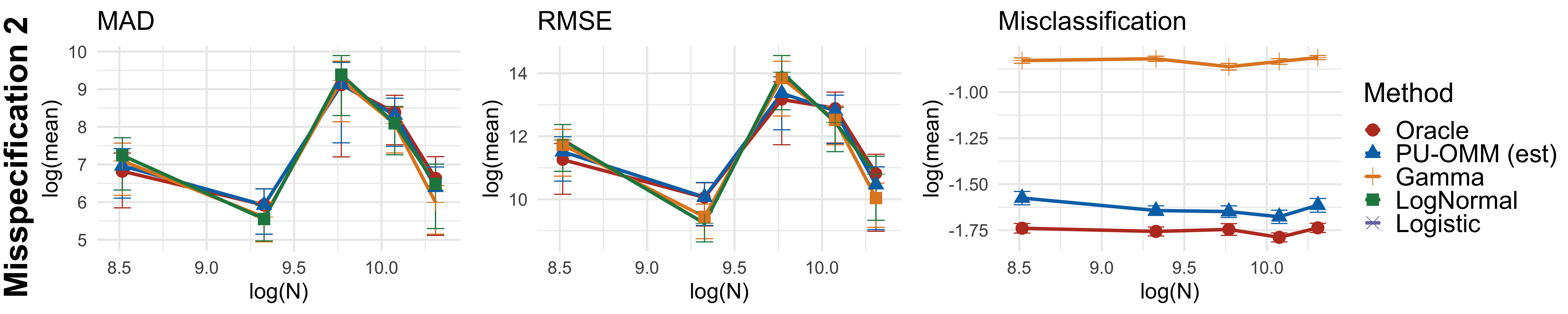}
  \label{fig:setting3-2}
\end{subfigure}
\caption{Prediction accuracy of Oracle, PU-OMM (est), PM-OMM (true), LogNormal-Logistic, Gamma-Logistic under Settings 1-3. For each row, the first two panels (MAD, RMSE) show prediction accuracy results using MAD (mean absolute deviation) and RMSE (root mean squared error). The last panel shows the misclassification rate of each method in predicting event occurrence. PU-OMM (true) refers to the PU-OMM method where true $\lambda_\epsilon$ (the value that was used in data generating process) is used and PU-OMM (est) refers to the PU-OMM method where $\lambda_\epsilon$ is chosen based on the goodness of fit of the observed occurrence. PU-OMM(true) is dropped in Setting 3 (Misspecification 2) because the true data generating process does not involve $\lambda_\epsilon$. We see that in terms of predicting occurrence of an event, PU-OMM performs better than the logistic model within the LogNormal-Logistic and Gamma-Logistic models, and performs similarly to the oracle method. In terms of predicting the size of each event,  statistically meaningful comparison between the considered methods is challenging, as the associated error bars overlap. Since MAD and RMSE evaluate prediction errors on an absolute scale, the performances were heavily affected by a small number of observations with large magnitudes (which arise since the model for an event size is multiplicative). In all cases, the performances of the PU-OMM methods were very similar to the oracle method. }\label{fig:evalModels-2}
\end{figure*}

In this section, we present supplementary figures for Section 4. Figure \ref{fig:evalModels} plots parametric estimation and prediction accuracy of each method including a PU-OMM method fitted with true $\lambda_\epsilon$. Figure \ref{fig:evalModels-2} plots prediction accuracy (for sizes and occurrences of events) of each method based on MAD, RMSE, and misclassification rate.

\subsection{Parametric estimation and prediction accuracy of each method including a PU-OMM method fitted with the true hyperparameter }

\newpage
\subsection{Prediction accuracy of each method using metrics of MAD, RMSE, and misclassification rate}

\end{document}